\documentclass{article}

\usepackage{hyperref}

\usepackage[accepted]{icml2024}

\usepackage[utf8]{inputenc} %
\usepackage[T1]{fontenc}    %
\usepackage{hyperref}       %
\usepackage{amsmath}
\usepackage[capitalize,noabbrev]{cleveref}
\usepackage{url}            %
\usepackage{booktabs}       %
\usepackage{amsfonts}       %
\usepackage{nicefrac}       %
\usepackage{microtype}      %
\usepackage{adjustbox}
\usepackage{multirow}
\usepackage{pifont}
\usepackage{soul}
\usepackage{nicefrac}
\usepackage{relsize}
\usepackage{enumitem}
\usepackage{xspace}
\usepackage{subcaption}
\usepackage{chngcntr}
\usepackage{wrapfig}
\usepackage{makecell}
\usepackage[skip=2pt plus1pt, indent=0pt]{parskip}
\usepackage{cleveref}
\usepackage{float}

\crefname{equation}{Eq.}{Eqs.}
\Crefname{equation}{Eq.}{Eqs.}

\DeclareMathAlphabet{\mathmybb}{U}{bbold}{m}{n}

\DeclareMathOperator*{\argmin}{arg\,min}

\newcommand{\R}{\mathbb{R}}
\newcommand{\E}{\mathbb{E}}

\newcommand\norm[1]{\left\lVert#1\right\rVert}

\newcommand{\xmark}{\ding{55}}%

\newcommand{\rr}{\textsc{RR}\xspace}

\newcommand{\fedrrr}{\textsc{Fed3R}\xspace}
\newcommand{\fedrrrrf}{\textsc{Fed3R-RF}\xspace}
\newcommand{\fedrrrft}{\textsc{Fed3R+FT}\xspace}
\newcommand{\fedrrrftf}{\textsc{Fed3R+FTfeat}\xspace}
\newcommand{\fedrrrftc}{\textsc{Fed3R+FTlp}\xspace}
\newcommand{\fedncm}{\textsc{FedNCM}\xspace}
\newcommand{\inaturalist}{iNaturalist-Users-120K\xspace}
\newcommand{\inat}{iNaturalist\xspace}
\newcommand{\landmark}{Landmark-Users-160K\xspace}
\newcommand{\gld}{Landmarks\xspace}
\newcommand{\cifar}[1] {\ifx\relax#1\relax Cifar100\xspace \else Cifar100-$\alpha$=#1\xspace \fi}

\newcommand{\ie}{\textit{i}.\textit{e}., }

\definecolor{darkblue}{RGB}{0,0,210}

\icmltitlerunning{Accelerating Heterogeneous Federated Learning with Closed-form Classifiers}

\begin{document}

\twocolumn[
\icmltitle{Accelerating Heterogeneous Federated Learning with Closed-form Classifiers}

\begin{icmlauthorlist}

\icmlauthor{Eros Fanì}{polito}
\icmlauthor{Raffaello Camoriano}{polito,xxx}
\icmlauthor{Barbara Caputo}{polito,zzz}
\icmlauthor{Marco Ciccone}{polito}

\end{icmlauthorlist}

\icmlaffiliation{polito}{Department of Computing and Control Engineering, Polytechnic University of Turin, Italy}
\icmlaffiliation{xxx}{Istituto Italiano di Tecnologia, Genoa, Italy}
\icmlaffiliation{zzz}{CINI Consortium, Rome, Italy}

\icmlcorrespondingauthor{Eros Fanì}{eros.fani@polito.it}

\icmlkeywords{Federated Learning, Ridge Regression, Random Features, Statistical Heterogeneity, Client Drift, Classifier Bias, Destructive Interference, Pre-trained models}

\vskip 0.3in
]

\printAffiliationsAndNotice{}  %

\begin{abstract}
\vspace{1em}
Federated Learning (FL) methods often struggle in highly statistically heterogeneous settings. Indeed, non-IID data distributions cause client drift and biased local solutions, particularly pronounced in the final classification layer, negatively impacting convergence speed and accuracy. To address this issue, we introduce \textit{Federated Recursive Ridge Regression} (\fedrrr). Our method fits a Ridge Regression classifier computed in closed form leveraging pre-trained features. \fedrrr is {immune to statistical heterogeneity} and is {invariant to the sampling order of the clients}. Therefore, it proves particularly effective in cross-device scenarios. Furthermore, it is fast and efficient in terms of communication and computation costs, requiring up to two orders of magnitude fewer resources than the competitors. Finally, we propose to leverage the \fedrrr parameters as an initialization for a softmax classifier and subsequently fine-tune the model using any FL algorithm (\fedrrr with Fine-Tuning, \fedrrrft). Our findings also indicate that maintaining a fixed classifier aids in stabilizing the training and learning more discriminative features in cross-device settings. Official website:  \href{https://fed-3r.github.io/}{https://fed-3r.github.io/}.

\end{abstract}

\section{Introduction}
\label{sec:intro}

Federated Learning (FL)~\cite{mcmahan2017communication} provides a practical framework for training machine learning models collaboratively across distributed \textit{clients} while ensuring privacy.
This decentralized approach involves multiple communication rounds between clients and a central server. During each round, clients leverage their private data to improve their local models. Then, they send the model updates to the server, which aggregates them and transmits the improved model to the next set of clients for further improvement.

While appealing, limiting the optimization on the client side presents several challenges. In real-world scenarios, billions of clients might be involved \cite{kairouz2021advances}, and data are often collected based on user preferences \cite{tan2022towards}, availability \cite{gu2021fast}, geographical location \cite{hsu2020federated,fantauzzo2022feddrive}, or personal habits \cite{fallah2020personalized,yang2018applied}.
This leads to data distributions across clients with inherent \textit{statistical heterogeneity} in the form of \textit{quantity skewness} \cite{li2019convergence, wang2020tackling, hsu2020federated}, \textit{label skewness}~\cite{karimireddy2020Scaffold, li2022federated, caldarola2022improving, feddrivev2}, or \textit{domain shift}~\cite{fantauzzo2022feddrive, nguyen2022fedsr, liu2021feddg}. 

As a result, training models that generalize well across the global underlying data distribution presents a major challenge. 
Specifically, convergence speed is hampered due to clients' sparse sampling and partial participation in successive rounds~\cite{li2019convergence, karimireddy2020mime}.
Furthermore, the process of aggregating model updates becomes particularly challenging in strongly heterogeneous settings.
This difficulty arises because biased local updates from individual clients can potentially steer the model away from global minimizers ~\cite{karimireddy2020Scaffold, li2020federated_fedprox, acar2021federated}.

Most of the approaches addressing such issues focus on regularizing the local objective to reduce model parameters drift~\cite{li2020federated_fedprox,karimireddy2020Scaffold,acar2021federated,ozfatura2021fedadc} or leveraging momentum to incorporate knowledge from previous updates and align the local optimization to the global direction~\cite{karimireddy2020mime,xu2021fedcm,kim2022communication,liu2023enhance}.

In particular, ~\citet{luo2021no} shows that model parameter drift in FL mainly involves neural network prediction heads. Indeed, deeper layers tend to be more susceptible to bias towards the individual client data distributions, while initial layers maintain better consistency in terms of representation similarity. As mainly studied in other areas such as Continual Learning~\cite{ratcliff1990connectionist, mccloskey1989catastrophic}, in classification this phenomenon occurs due to the inherent nature of the softmax classifier, which is prone to forgetting if updated by sampling data in a non-i.i.d. or class-imbalanced manner~\cite{kirkpatrick2017overcoming}, as the most recently acquired knowledge tends to be more relevant than the older one, resulting in \textit{recency bias} (or \emph{catastrophic forgetting}) \cite{mai2021supervised, masana2022class, wu2019large, lyu2024overcoming}. Similarly, in FL, the local optimization biases the classifier towards the local distribution, which can result in overwriting past clients' knowledge~\cite{legate2023re, caldarola2022improving}. %
This problem is exacerbated in realistic \emph{cross-device} scenarios with large number of devices, where clients may not be revisited during training \cite{ruan2021towards}, making the optimization slower and unstable.

To address this issue, we propose \textit{Federated Recursive Ridge Regression} (\fedrrr), a novel approach to FL 
leveraging pre-trained representations to train classifiers that are immune to statistical heterogeneity by design. %
The \fedrrr classifier can be efficiently trained and incrementally updated in closed form by repurposing the Ridge Regression (\rr) online formulation for FL. Each client computes its local RR statistics using the feature maps generated by the pre-trained feature extractor and sends them to the server, where they are aggregated and used to compute the \rr classifier. The \fedrrr solution is equivalent to the centralized \rr solution and invariant to the sampling order of the clients. Our strategy allows aggregating client models exactly, efficiently, and without the need for backpropagation. Moreover, each client necessitates only a single round of communication with the server, contrary to traditional gradient-based methods. %

Furthermore, to address non-linearities of the latent space and the distribution shift of the target task from the pre-trained representation, we present \fedrrr with Random Features (\fedrrrrf), a kernelized version of \fedrrr, and \fedrrr with Fine-Tuning (\fedrrrft), which can update both feature extractor and the classifier jointly.

Finally, we repurpose \rr as a tool to quantitatively assess the quality of the feature extractors. Indeed, learning an \rr classifier on the features of a trained feature extractor allows decoupling the contributions of the feature extractor and the classifier on the final performance. With this, we find that fine-tuning only the feature extractor while keeping the \fedrrr classifier fixed not only helps to counteract client drift and destructive interference during the aggregation phase but also improves the quality of the features in settings with strong statistical heterogeneity.

\paragraph{Contributions} %
\begin{itemize}
    \item We propose \fedrrr, a federated version of Ridge Regression, to efficiently learn a linear classifier that is immune to statistical heterogeneity. We also propose \fedrrrrf, a kernelized version of the algorithm based on random features to handle the non-linearities of the input space. %
    \item We demonstrate that \fedrrr significantly accelerates training, converging faster than FL other methods. Additionally, we show that \fedrrr reduces communication and computational costs by up to two orders of magnitude compared to other methods.
    \item We show that \fedrrr can also be employed as a classifier initialization for fine-tuning representations and how it can stabilize the training in highly heterogeneous settings. We evaluate the effectiveness of our proposed algorithms on the \gld and \inat datasets~\cite{hsu2020federated}, two realistic and cross-device FL scenarios for visual classification with thousands of clients and classes.
    \item We show how to repurpose \rr as a tool to discern the contributions of the feature extractor and the classifier to the final model performance. We find that fine-tuning the feature extractor while keeping the \fedrrr classifier fixed greatly improves the features' quality, robustness to client drift, and destructive interference. 
\end{itemize}

\section{Related Works}
\label{sec:related}

\paragraph{Statistical heterogeneity in FL.}
Despite the effectiveness of current FL approaches in homogeneous scenarios with i.i.d. data, addressing statistically heterogeneous and realistic settings remains challenging \cite{kairouz2021advances}. Private data exhibits biases due to factors such as personal habits and geographical locations \cite{hsu2020federated, kairouz2021advances, fallah2020personalized}, causing variations among clients in categories, domains, and dataset sizes. This bias induces \textit{client drift} \cite{karimireddy2020Scaffold}, leading local models to converge toward different minima, deviating from the global direction, resulting in noisy, unstable learning trends~\cite{li2019convergence, caldarola2022improving}. %

\paragraph{Optimization-based methods for heterogeneous FL.}
To reduce the impact of heterogeneity, simple solutions involve limiting the drift of the local models with regularization techniques. For instance, FedProx~\cite{li2020federated_fedprox} introduces a penalization in the local objectives to prevent divergence from the global model. Other methods, such as Scaffold \cite{karimireddy2020Scaffold}, leverage stochastic variance reduction \cite{reddi2016stochastic} to correct the local direction with the global one. FedDyn~\cite{acar2021federated} aligns local and global stationary points at convergence, enjoying the same convergence properties of Scaffold.
Still, its practical effectiveness in cross-device settings is limited since it is often prone to parameter explosion~\cite{varno2022adabest}. 

Other approaches aim to reduce client drift by exploiting the history of previous updates, incorporated with momentum or adaptive optimizers~\cite{wang2019slowmo,reddi2020adaptive}. In particular, FedAvgM~\cite{hsu2019measuring} employs momentum in the server-side aggregation, demonstrating effectiveness in realistic settings~\cite{hsu2020federated}. Other works introduce client-side momentum~\cite{kim2022communication, xu2021fedcm, karimireddy2020mime} to guide the local updates in the direction followed by the global model. Finally, Mime \cite{karimireddy2020mime} aims at replicating the behavior of models trained on i.i.d. data by combining stochastic variance reduction and client-side momentum. 

Despite these methods being theoretically principled, our work empirically reveals their inherent instabilities in real-world cross-device FL scenarios. In contrast, we demonstrate that training classifiers with closed-form solutions and exact aggregation can be dramatically faster and communication efficient than gradient-based optimization in practical cross-device scenarios. Additionally, we illustrate that integrating our method with FL optimizers can further expedite convergence through a final fine-tuning stage.

\paragraph{Classifier bias and destructive interference.}
A natural direction to study the effect of heterogeneity is to analyze its impact on different parts of the model. On this matter, \citet{luo2021no} showed that the bias of the clients towards local data distributions is significantly more pronounced in the deeper layers, with a peak in the last one, \ie the prediction head. This phenomenon has also been observed in Continual Learning on sequences of heterogeneous task distributions~\cite{ramasesh2020anatomy, davari2022probing, kim2023stability}. Other works observed that biased classifiers and misaligned features create a vicious cycle \cite{zhou2022fedfa, li2023no}. On one side, discriminative features are needed to train models effectively, as convergence speed improves when gradients are more aligned~\cite {nguyen2022begin}. However, heterogeneity heavily affects the prediction head, which suffers from destructive interference during the aggregation phase and hampers the features learning process. %
Indeed, \citet{yu2022tct} shows that learning the feature extractor and the classifier in two separate phases may be beneficial in FL, as also observed in other fields ~\cite{kang2019decoupling, wang2022schedule}.

Previous research attempted to mitigate classifier biases at the end of FL training by retraining only the classifier on the server with generated virtual features~\cite{luo2021no, shang2022federated, nguyen2022begin}. %
However, this approach remains sub-optimal as it is based on the quality of the feature representation and generative process, which could negatively affect the retrained classifiers. 
More recently, \citet{li2023no} proposed a fixed synthetic classifier, motivated by the simplex geometry of the logits space induced by the neural collapse~\cite{papyan2020prevalence}.

In this work, we take a different turn and tackle the classifier bias problem with a principled approach. 
Starting from a pre-trained representation, we employ a \textit{one-vs-rest} Ridge Regression (RR) classifier that can be trained 
in a distributed setting with exact aggregation. This efficiency stems from an online formulation that recovers the closed-form solution of the centralized problem by effectively providing a classifier that does not suffer from statistical heterogeneity and is invariant to the actual federated split. \citet{legate2023guiding} adopts a similar rationale for a Nearest Class Mean (NCM) classifier, avoiding gradient updates using class centroids. While NCM may be effective on simpler datasets, we demonstrate its weakness in realistic scenarios, contrasting the consistent performance of RR.

We refer to \cref{sec:more_related} for additional related works on existing \rr-based methods in distributed learning and vertical FL and a broader overview of the existing literature on transfer learning methods with pre-trained models in FL.

\section{Background}
In this section, we provide a concise overview of the FL framework and the fundamental concepts of Ridge Regression before formally describing our algorithm.

\subsection{FL Problem Formulation}
Let $\mathcal{K}$ be the set of all the clients involved in the training with cardinality $|\mathcal{K}| = K$, and let $\mathcal{S}$ be the server that orchestrates the training procedure. 
Each client $k \in \mathcal{K}$ has access to a private local dataset $\mathcal{D}_k$ of size $n_k = |\mathcal{D}_k|$; neither the server nor the other clients can access $\mathcal{D}_k$. 
Each local dataset $\mathcal{D}_k$ is composed of $n_k$ pairs $(x, y) \sim P_k$, where $x \in \mathcal{X}, y \in \mathcal{Y}$. 
Here, $\mathcal{X}$ and $\mathcal{Y}$ represent the input and output spaces, and $P_k$ is the joint data distribution associated with client $k$. 

The global federated objective is given by:
\begin{equation}
    \theta^* = \argmin_\theta \sum_{k\in\mathcal{K}} \mathcal{L}_k(\mathcal{M}; \mathcal{D}_k),
\end{equation}
where $\mathcal{L}_k = \sum_{(x, y)\in \mathcal{D}_k}\ell(\mathcal{M}(x; \theta), y)$  is the Local Empirical Risk associated to the client $k$, computed according to a loss function $\ell$ (e.g., cross-entropy), and $\mathcal{M}$ is a model parameterized by $\theta$.
At each round $t$, a subset of selected clients $\mathcal{K}' \subseteq \mathcal{K}$ receives the global parameters $\theta^{t-1}$ of the previous round from the server, initializes the local parameters $\theta_k = \theta^{t-1}$ and optimizes them using the private datasets $\mathcal{D}_k$, obtaining the new parameters $\theta^t_k$.
Then, the locally optimized model parameters $\theta_k^t$ are shared with the server $\mathcal{S}$, which aggregates them according to the specific FL algorithm. For instance, the FedAvg \cite{mcmahan2017communication} aggregation rule is a weighted average of clients' models $ \theta^t = \sum_{k \in \mathcal{K}'} \frac{n_k}{n} \theta_k^t$, where $n = \sum_{k \in \mathcal{K}'} n_k$. The server broadcasts the aggregated model $\theta^t$ to the new active clients. The process is repeated for several rounds until convergence.

\subsection{Closed-form Ridge Regression (\rr)}
\label{sec:brr}
Our work is based on the idea of using one-vs-rest classifiers such as \textit{least-squares regressors}~\cite{stigler1981gauss, bjorck1996numerical,rifkin2003regularized} that admit a closed-form solution and can be computed efficiently.
We first define the problem for the \textit{centralized setting}, where samples from a dataset $\mathcal{D}$ can be accessed simultaneously.

Although simple least-squares empirical risk minimization is generally prone to overfitting~\cite{bishop2006pattern}, it can easily be augmented with $\mathcal{L}_2$ regularization (controlled by a Tikhonov hyper-parameter $\lambda \in \mathbb{R}^+$), obtaining a Ridge Regression~\cite{boyd2004convex} 
problem:
\begin{equation}
    \label{eq:brr_linear_obj}
    W^* = \argmin_{W \in \R^{p \times C}} \norm {Y - XW}^2 + \lambda \norm{W}^2,
\end{equation}
where $X \in \mathbb{R}^{n \times p}$ is the matrix of the $n$ stacked input samples, $Y \in \mathbb{R}^{n \times C}$ is the matrix of the stacked one-hot-encoding vectors of the corresponding $C$ classes, and $p$ is the input dimensionality. The solution $W^*$ constitutes the optimal parameters for the linear predictor $f(x; W) = W^\top x$.

The problem in \cref{eq:brr_linear_obj} admits a closed-form solution:
\begin{equation}
    \label{eq:rr_solution}
    W^{*} = (X^{\top} X + \lambda I_p)^{-1} X^{\top} Y,
\end{equation}
where $I_p$ is the $p \times p$ identity matrix.

Since $X^\top X + \lambda I_p \succ 0$ for any $\lambda > 0$, no additional assumptions on the rank or the dimensions of the matrix $X$ are required to prove that the optimal solution $W^*$ exists \cite{boyd2004convex}. Moreover, \rr can be directly applied to classification, as introduced in \citet{rifkin2003regularized}, and it converges in probability to the optimal Bayes classifier as $n$ tends to infinity \cite{steinwart2008support, bartlett2006convexity, krr2}.

\subsection{Handling Non-linear Input Spaces in \rr}
While simple and powerful, \rr is a linear classifier whose performance is tied to the separability of the input space.
To handle the non-linearities of the input space, we map $\mathcal{X}$ onto a latent feature space $\mathcal{Z} \subseteq \mathbb{R}^d$ using a pre-trained feature extractor $\varphi: \mathcal{X} \rightarrow \mathcal{Z}$ and apply \cref{eq:rr_solution} directly on the feature maps to obtain the optimal predictor.

For clarity, we express \cref{eq:rr_solution} for a linear classifier $W^* \in \mathbb{R}^{d \times C} $ whose input space is $\mathcal{Z}$, as:
\begin{equation}
\label{eq:rr_sol_better}
    W^* = (A + \lambda I_d)^{-1} b
\end{equation}
Here, $A := Z^\top Z = \sum_{(x, y) \in \mathcal{D}} \varphi(x)\varphi(x)^\top \in \mathbb{R}^{d \times d}$ is the covariance matrix of samples in the mapped space, where $Z \in \mathbb{R}^{n \times d}$ is the matrix of mapped input samples with each row $Z_i = \varphi(X_i)$. Also, $b := Z^\top Y = \sum_{(x, y) \in \mathcal{D}} \varphi(x)e_y^\top  \in \mathbb{R}^{d \times C}$, and $e_y$ is the one-hot encoding vector for class $y$.

Although a pre-trained feature extractor $\varphi$ can be used for handling non-linearities in the input space, the performance depends on the quality of $\varphi$ on the target task. To further improve the latent space's separability, we also consider employing Kernel Ridge Regression (KRR) in our method. KRR is a nonparametric learning algorithm that uses kernel functions to implicitly address the non-linearity of the input space \cite{krr, krr2}. However, the kernel matrix's space complexity is $\mathcal{O}(n^2)$, in contrast to the covariance matrix $A$ with space complexity of $\mathcal{O}(d^2)$. 
For sizable datasets ($n \gg d$), storing the kernel matrix and computing the exact KRR solution becomes impractical.
To overcome this bottleneck, we employ Random Features KRR \cite{random_features}, a data-independent subsampling scheme enabling optimal generalization properties while reducing the computational complexity of KRR~\cite{rudi2017generalization} through an approximate nonlinear mapping of the input features.
Its properties are particularly suitable for the FL setting, 
enabling us to keep the same formulation as our algorithm's linear version, as shown in the next section.

\section{Method}
\label{sec:method}

We now present Federated Recursive Ridge Regression (\fedrrr), and show how recursive least squares can be elegantly repurposed to the FL setting. 
Each client contributes to the $A$ and $b$ matrices of \cref{eq:rr_sol_better} by independently computing local statistics, which are then collected and aggregated by the server and used to compute the global RR classifier in closed form.
Moreover, we introduce \fedrrrrf, a kernelized version of our algorithm that uses random features to approximate the KRR solution.

\subsection{Federated Recursive Ridge Regression (\fedrrr)}
\label{sec:fed3r}

While the least-squares problem can be effectively solved in a closed form via \cref{eq:rr_solution}, in principle it needs access to the entire dataset $\mathcal{D}$, which is not always available if data is accessed sequentially ~\cite{camoriano2017incremental,wang2022schedule}
or is distributed across devices, as in FL.

Luckily, thanks to the linearity of \cref{eq:rr_solution,eq:rr_sol_better}, when new samples become available, the optimal solution can be exactly updated by recursive least squares~\cite{kailath2000linear}. This method computes solutions recursively and efficiently via Sherman-Morrison-Woodbury or Cholesky updates~\cite{sherman1950adjustment, hager1989updating}.

Alternatively, the \rr statistics $A$ and $b$ can be cumulatively updated without re-computing them from scratch on the entire dataset. The solution can then be computed using the updated statistics by solving a linear system. Our method is based on the observation that the matrices $A$ and $b$ can be incrementally computed, for instance, by simply summing over the samples of the dataset $\mathcal{D}$. Exact and efficient incremental \rr updates enable several continual and incremental classification methods \cite{camoriano2017incremental, wang2022schedule}, as RR admits an equivalent exact incremental solution \cite{bjorck1996numerical, sayed2008} that we reformulate specifically for the FL context, finally leading to our \fedrrr algorithm.

In practice, thanks to the associative property of the sum, we can break the matrices $A$ and $b$ into the contributions of the clients' local datasets $\mathcal{D}_k$:
\begin{align}
    \label{eq:A}
    A &= \sum_{(x, y) \in \mathcal{D}} \varphi(x)\varphi(x)^\top = \sum_{k\in\mathcal{K}} \sum_{(x, y) \in \mathcal{D}_k} \varphi(x)\varphi(x)^\top \nonumber \\
      &= \sum_{k \in \mathcal{K}} Z_k^\top Z_k = \sum_{k \in \mathcal{K}} A_k,
\end{align}
\begin{align}
    \label{eq:b}
    b &= \sum_{(x, y) \in \mathcal{D}} \varphi(x)e_y^\top = \sum_{k\in\mathcal{K}} \sum_{(x, y) \in \mathcal{D}_k} \varphi(x)e_y^\top \nonumber \\
      &= \sum_{k \in \mathcal{K}} Z_k^\top Y_k = \sum_{k \in \mathcal{K}} b_k.
\end{align}

This is true for all the possible partitions of the underlying dataset $\mathcal{D}$ such that $\mathcal{D} = \bigcup_{k \in \mathcal{K}}\mathcal{D}_k$ is the union of the local datasets $\mathcal{D}_k$, with $\mathcal{K}$ the set of all the clients.

In \fedrrr, each client $k$ computes its local $A_k$ and $b_k$ statistics and shares them with the server, where they are aggregated and employed to calculate $W^*$. Hence, the \fedrrr solution is mathematically equivalent to the centralized \rr solution, independently of the federated split. Therefore, it inherits all the generalization properties of \rr. In particular, it achieves optimal convergence rates in probability \cite{caponnetto2007optimal}. 

Finally, to address possible class unbalanced distributions over $\mathcal{D}$, we normalize $W^*$ by dividing each column by its class norm, similar to the approach used by the authors of \cite{legate2023guiding}: $W^*_c \leftarrow W_c^* / \norm{W_c^*}$. %

\subsection{\fedrrr with Random Features (\fedrrrrf)}
\label{sec:fed3rrf}

As pre-trained feature extractors may not be expressive enough to separate features for complex learning problems linearly, we also introduce \fedrrrrf, which first performs a nonlinear random features mapping of the latent feature space to a new $D$-dimensional feature space by approximating the corresponding kernel feature map, where $D > d$ is a hyper-parameter. Consequently, all the dimensionalities of the statistics that depended on $d$ here depend on $D$.
\fedrrr and \fedrrrrf are summarized in \cref{alg:fed3r}.

\begin{algorithm}[tb]

   \caption{- \textbf{\fedrrr} and \textcolor{darkblue}{\textbf{\fedrrrrf}} %
   }
   \label{alg:fed3r}
\begin{algorithmic}
    \STATE \textbf{Require:}
    \STATE Server $\mathcal{S}$, clients $\mathcal{K}$
    \STATE Fixed pre-trained feature extractor $\varphi: \mathcal{X} \rightarrow \R^d$
    \STATE \textcolor{darkblue}{Random features $\omega \in \R^{d \times D}$}
    \STATE Hyper-parameter $\lambda > 0$ %
    \STATE \textbf{Clients} $k \in \mathcal{K}$:
    \FOR{\textbf{each} client $k \in \mathcal{K}$ in parallel}
        \STATE $Z_k = \varphi(X_k)$
        \STATE \textcolor{darkblue}{Map $Z_k$ to a D-dimensional space using the RF $\omega$}
        \STATE $A_k = Z_k^\top Z_k$, \hspace{.5em} $b_k = Z_k^\top Y_k$
    \ENDFOR
    \STATE \textbf{Server} $\mathcal{S}$:
    \STATE Collect all the clients' statistics
    \STATE Compute $A = \sum_{k\in\mathcal{K}}A_k$, \hspace{.2em} $b = \sum_{k\in\mathcal{K}}b_k$
    \STATE Apply \cref{eq:rr_sol_better} to get $W^*$
    \STATE Normalize $W^*$: $W^*_c \leftarrow W^*_c / \norm{W^*_c} \hspace{.5em} \forall c \in [C]$

\end{algorithmic}
\end{algorithm}

\subsection{\fedrrr and \fedrrrrf Properties}
\label{sec:fed3rprops}

The \fedrrr (\fedrrrrf) solution computed using all the local datasets $\mathcal{D}_k$ is mathematically equivalent to the corresponding centralized \rr (\rr with Random Features) solution using $\mathcal{D}$. Consequently, the federated classifiers inherit all the properties and guarantees of the centralized ones. Additionally, both methods exhibit three fundamental and desirable properties related to the FL setting, which we list below. Finally, for a discussion on the privacy guarantees of our method, we refer the reader to \cref{sec:privacy}.

\paragraph{Immunity to statistical heterogeneity.} As \cref{eq:A,eq:b} show, due to the associative and commutative properties of the sum, once all the clients have shared their statistics with the server, the matrices $A$ and $b$ are the same for all possible partitions of the dataset. Hence, the \fedrrr solution is invariant to the particular data split across the clients; in other words, \fedrrr is immune to statistical heterogeneity and is invariant to the clients' sampling order. Consequently, \fedrrr guarantees the same final solution given any FL split of the same dataset $\mathcal{D}$.

\paragraph{Clients are sampled only once.} In \fedrrr, each client only needs to communicate its statistics once, meaning it only needs to be sampled once. If we assume that, as for classical FL algorithms, $\kappa$ clients are sampled during each round without replacement, \fedrrr requires exactly $\lceil K/\kappa \rceil$ rounds to converge to its final optimal solution, and no asymptotic convergence proof is required. The convergence is exact and guaranteed after $\lceil K/\kappa \rceil$ rounds. Therefore, the higher the participation rate, the faster \fedrrr converges. This is not generally guaranteed for gradient-based FL algorithms.

\paragraph{Differences with gradient-based FL algorithms.} Unlike gradient-based FL algorithms, \fedrrr does not rely on common assumptions such as the smoothness of clients' objectives or the unbiasedness and bounded variance of stochastic gradients \cite{kairouz2021advances, karimireddy2020mime, karimireddy2020Scaffold, acar2021federated}. In addition, \fedrrr does not require assuming bounded gradient dissimilarity among clients, which formalizes the effect of heterogeneous local datasets.

\subsection{\fedrrr with Fine-Tuning (\fedrrrft)}
\label{sec:fed3rft}
The proposed \fedrrr algorithm is a fast and efficient solution to learn a classifier with guarantees of being immune to statistical heterogeneity. However, \fedrrr performance relies on the quality of the pre-trained feature extractor, which is frozen. Similar to the approaches from \citet{wang2022schedule} and \citet{legate2023guiding}, we propose \fedrrr with Fine-Tuning (\fedrrrft), a two-stage algorithm where a fine-tuning stage follows the classifier initialization.

First, \fedrrrft learns a \fedrrr classifier using a pre-trained feature extractor. Then, it initializes a softmax classifier using the parameters of the \fedrrr classifier. Finally, the whole model is fine-tuned using a traditional FL algorithm. As the \fedrrr classifier is the optimal Regularized Least Squares classifier obtained using the pre-trained feature extractor, it provides a stable starting point that can mitigate client drift and destructive interference during aggregation.

However, due to the \fedrrr classifier's derivation from the mean squared loss, the entropy of its predictions distribution may not directly correspond to that of the cross-entropy (CE) loss employed in the fine-tuning phase. Consequently, the shape of the two loss landscapes can significantly vary. To solve this issue, we calibrate the entropy of the \fedrrr initialization by adjusting the temperature of the softmax function (see more details in \cref{sec:impl_det}).

We propose three different fine-tuning strategies for \fedrrrft. The first involves fine-tuning the entire model, which we actually refer to as \fedrrrft\footnote{We sometimes use \fedrrrft\textsc{feat+lp} for convenience, meaning that we fine-tune both feature extractor and classifier.}. In some cases, the pre-trained features may already be robust enough, so it is reasonable only to fine-tune the classifier. We call this variant \fedrrrftc. On the other hand, keeping the \fedrrr classifier constant while fine-tuning the feature extractor could help minimize destructive interference, especially in cross-device scenarios with high statistical heterogeneity, where the classifier is often the most affected layer \cite{luo2021no, li2023no}. We refer to this last variant as \fedrrrftf.

\section{Experiments}

\begin{figure}[t]
    \centering
    \includegraphics[width=\linewidth]{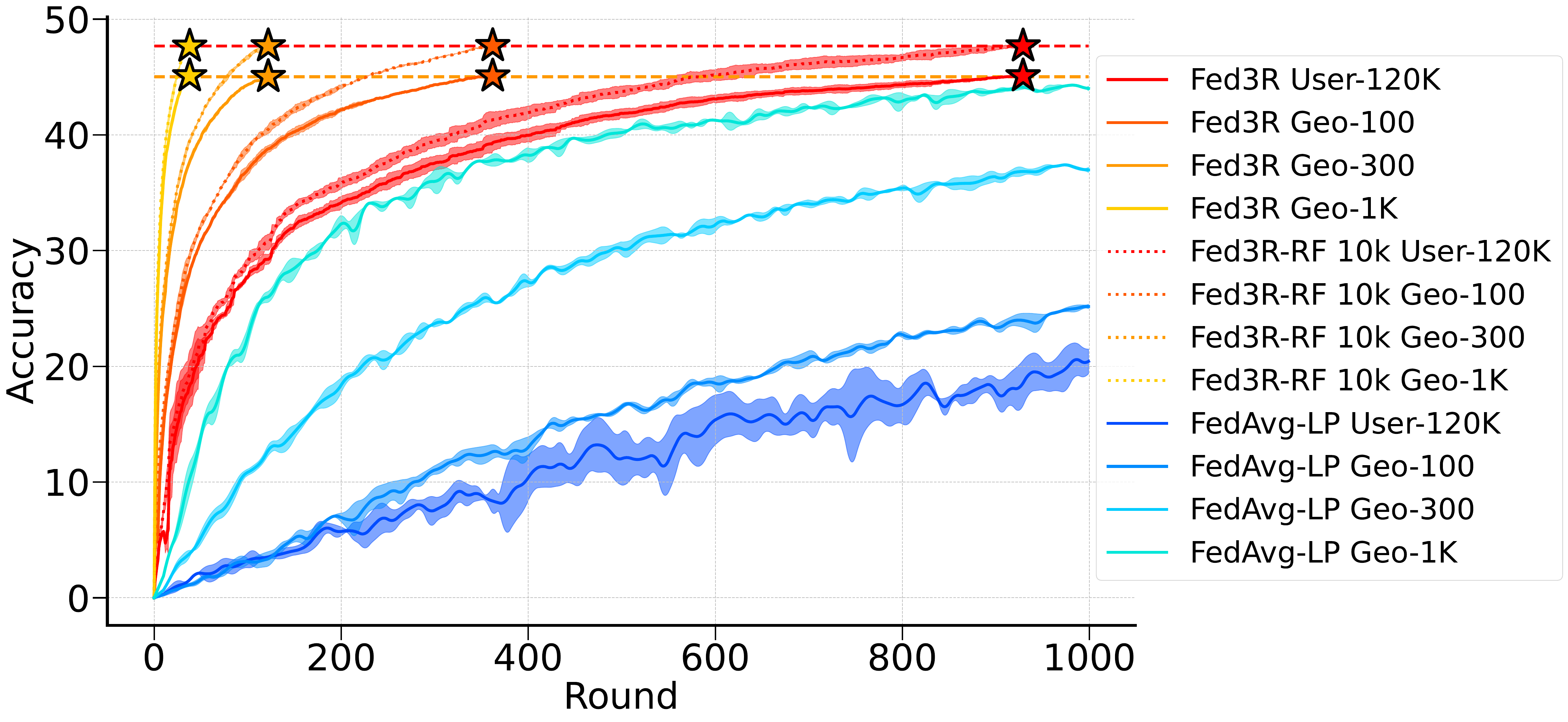}
    \caption{\fedrrr and \fedrrrrf invariance to different iNaturalist splits. All the curves converge to the same values, showing how both methods are immune to statistical heterogeneity. }
    \label{fig:stat_het_inat}
    \vspace{-1em}
\end{figure}

\begin{figure*}
    \centering
    \begin{subfigure}[b]{0.33\linewidth}
        \includegraphics[width=\linewidth]{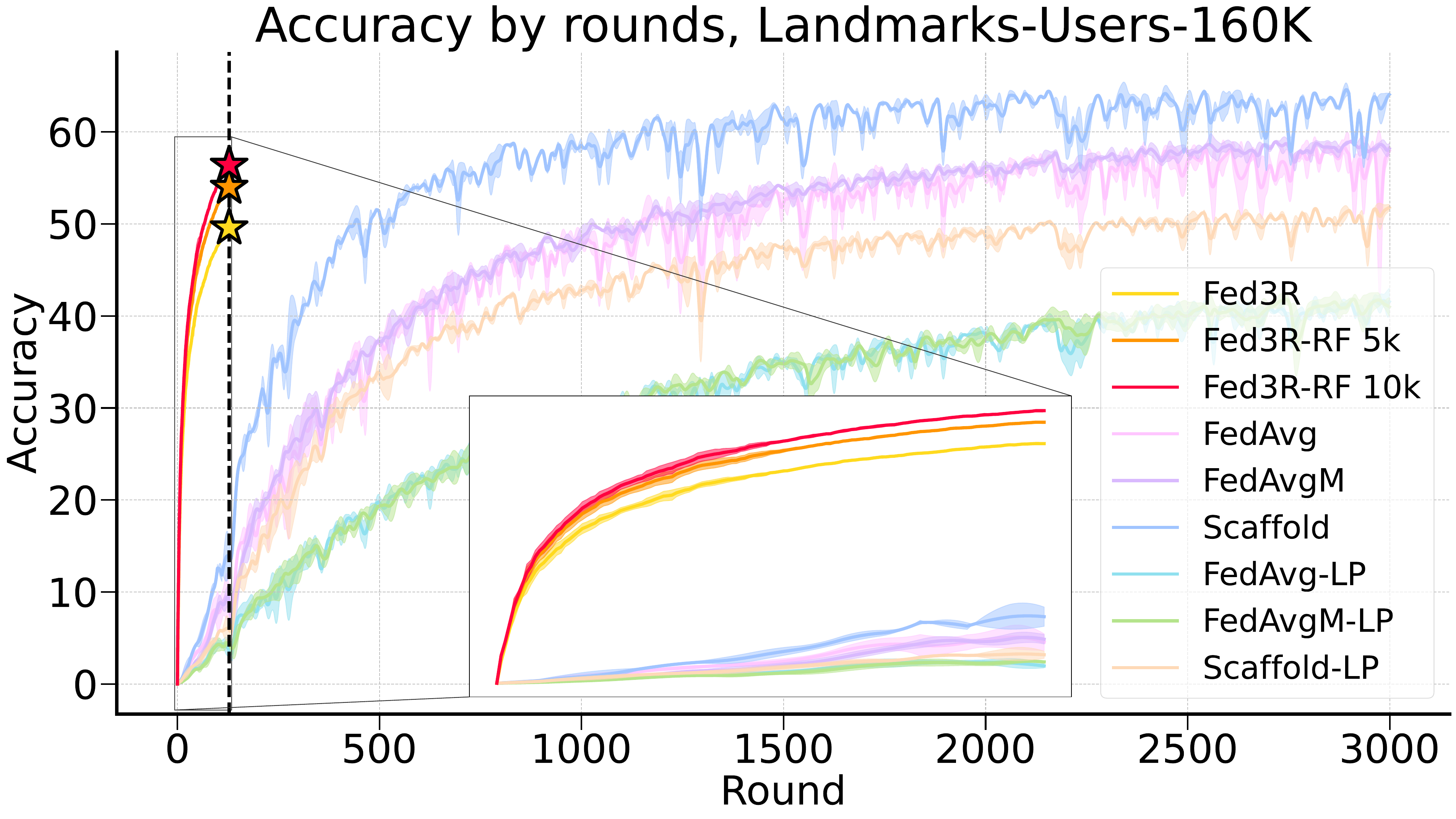}
    \end{subfigure}
    \begin{subfigure}[b]{0.33\linewidth}
        \includegraphics[width=\linewidth]{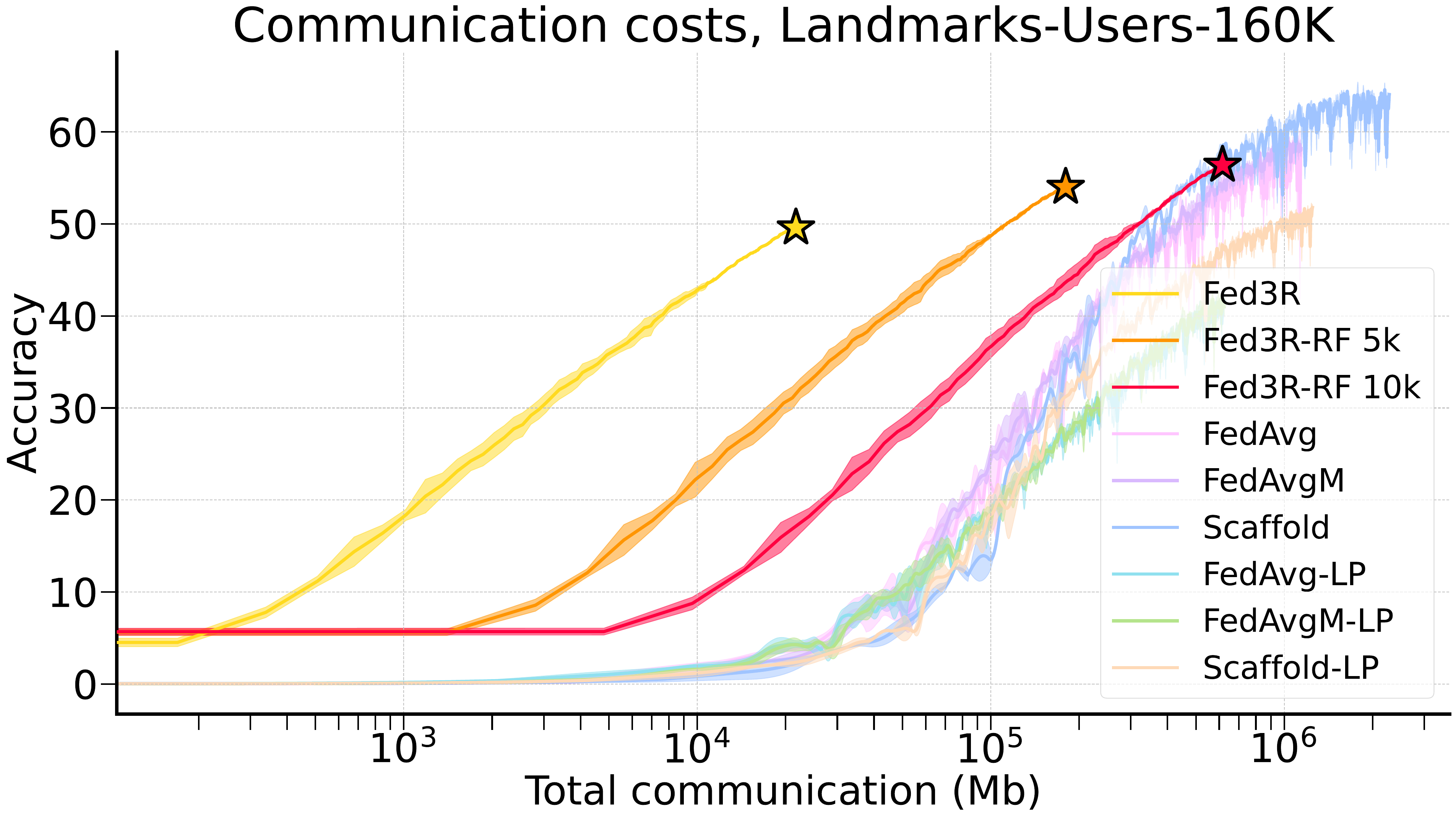}
    \end{subfigure}
    \begin{subfigure}[b]{0.33\linewidth}
        \includegraphics[width=\linewidth]{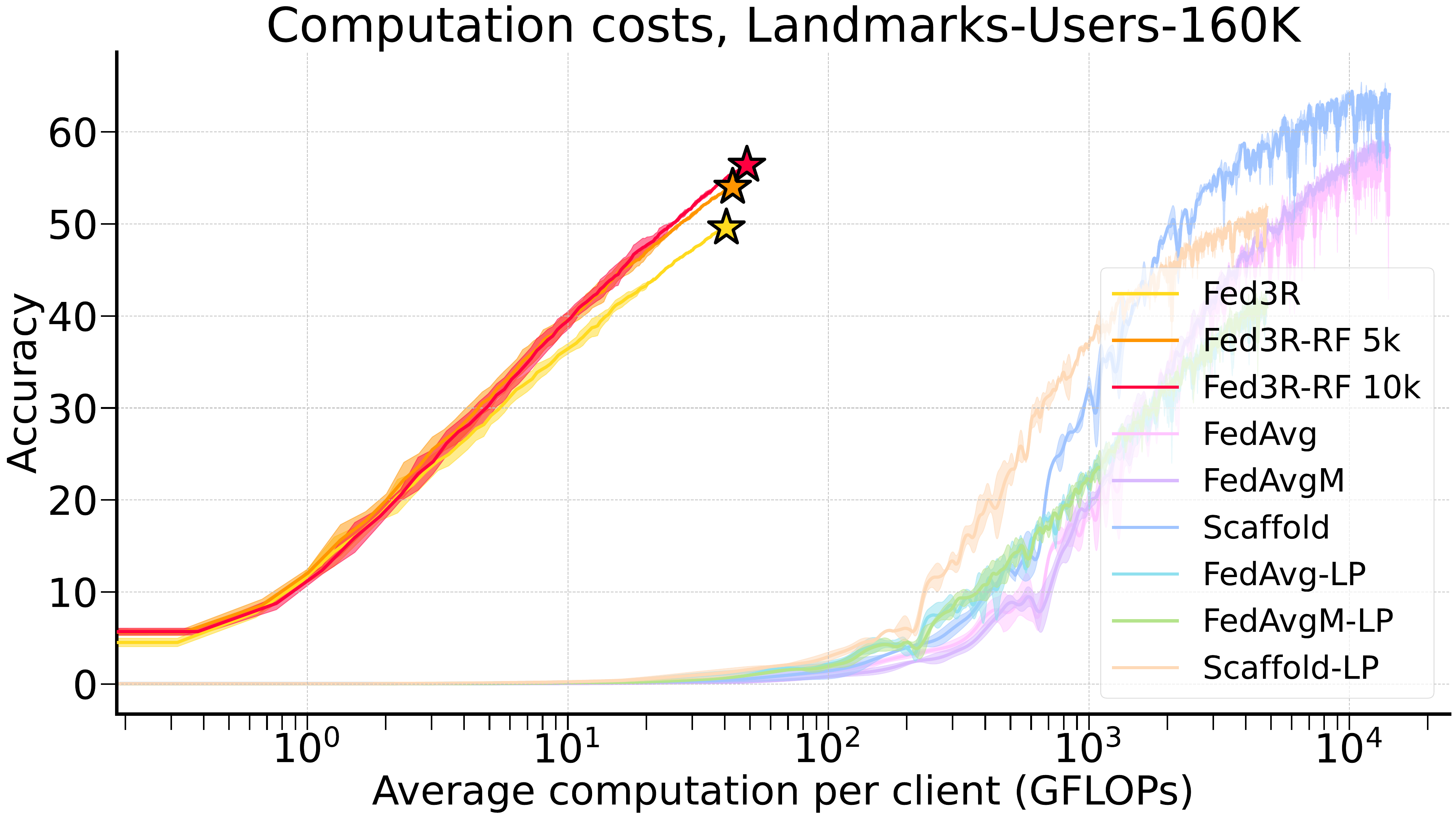}
    \end{subfigure}
    \begin{subfigure}[b]{0.33\linewidth}
        \includegraphics[width=\linewidth]{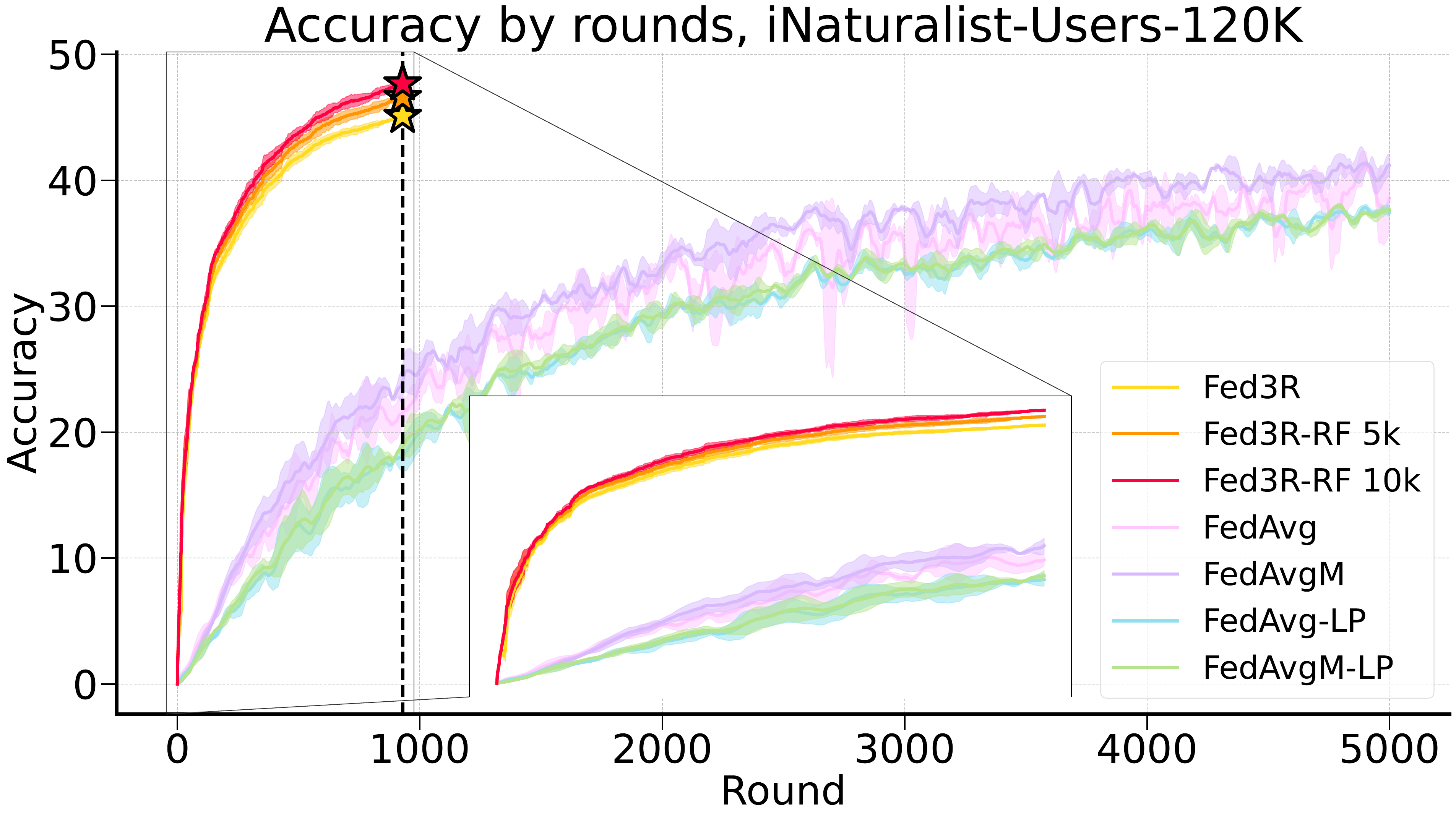}
    \end{subfigure}
    \begin{subfigure}[b]{0.33\linewidth}
        \includegraphics[width=\linewidth]{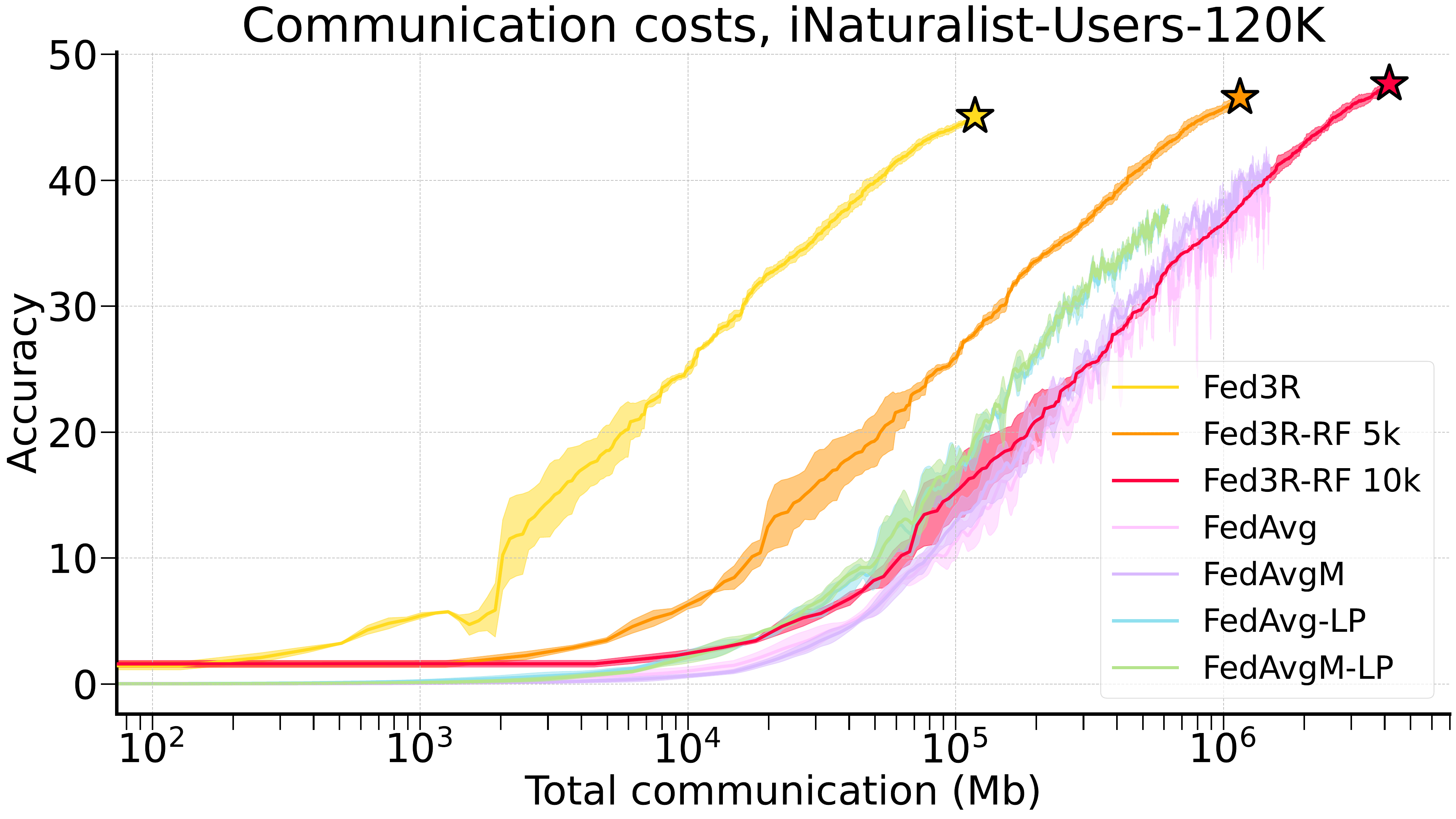}
    \end{subfigure}
    \begin{subfigure}[b]{0.33\linewidth}
        \includegraphics[width=\linewidth]{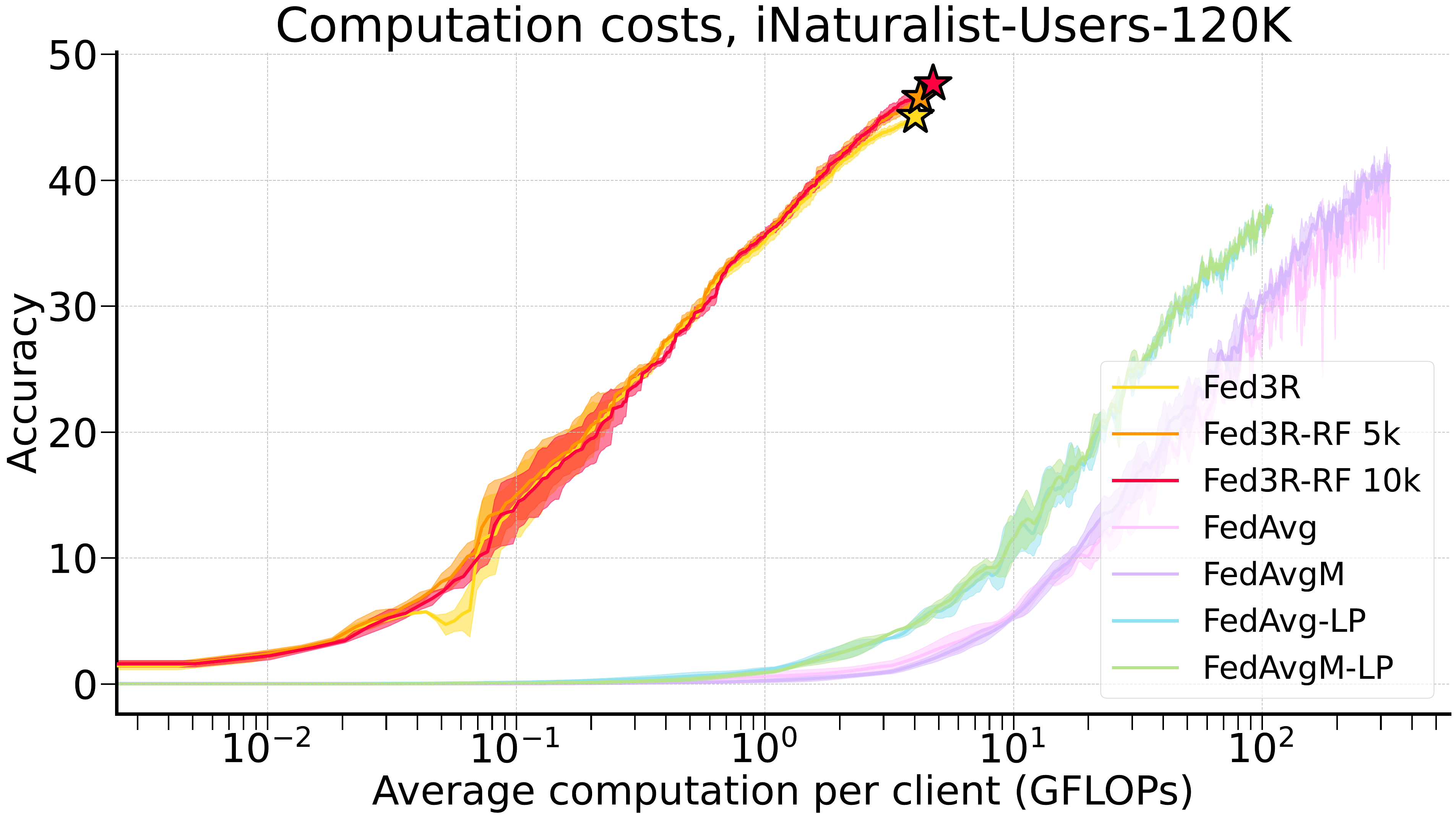}
    \end{subfigure}
    \caption{Comparison between \fedrrr and the baselines. From left to right: accuracy vs rounds, accuracy vs communication budget, accuracy vs average computation per client. Top row: \gld results, Bottom row: \inat results. Fed3R shows clear advantages regarding convergence speed, communication, and computation budget required.}
    \label{fig:fed3r}
\end{figure*}

In this section, we empirically evaluate the performances of our proposed methods in terms of accuracy, convergence speed, communication, and computational costs. First, we empirically show that \fedrrr and \fedrrrrf are both immune to statistical heterogeneity and their performances are equivalent to the ones of the corresponding centralized \rr solutions. Then, we compare \fedrrr and \fedrrrrf with FedAvg \cite{mcmahan2017communication}, FedAvgM~\cite{hsu2019measuring}, and Scaffold~\cite{karimireddy2020Scaffold}. Finally, we show how \fedrrr can effectively bootstrap training and use it as powerful initialization when combined with other optimization methods using \fedrrrft.

\paragraph{Datasets.} For the evaluation we choose two large-scale image classification datasets, \gld \cite{weyand2020google} and \inat \cite{van2018inaturalist}, both partitioned as proposed in \cite{hsu2020federated}~\footnote{For \gld and \inat, we always mean the \landmark and \inaturalist partition, respectively, except when differently declared.}.
These datasets emulate realistic FL scenarios, as they offer over $10^5$ training images and involve thousands of heterogeneous clients (see \cref{tab:datasets} in \cref{sec:impl_det} for additional details). %
We select $10$ clients per round in all our experiments, except when differently declared, simulating a participation rate of $\approx 0.8 \%$ for \gld and $\approx 0.1 \%$ for \inat.

\paragraph{Models and baselines.} All the experiments are conducted using a MobileNetV2 architecture \cite{mobilenetv2} pre-trained on ImageNet-1k \cite{deng2009imagenet}. As baselines we included FedAvg \cite{mcmahan2017communication}, FedAvgM \cite{hsu2019measuring} and Scaffold \cite{karimireddy2020Scaffold}. We do not include FedDyn \cite{acar2021federated}, Mime, and MimeLite \cite{karimireddy2020mime} because they fail to converge in most of the \fedrrrft setting. Moreover, we do not include Scaffold in all the \inat experiments, as it fails to converge. We refer to \cref{sec:impl_det} for additional implementation details. 

\paragraph{Additional details.} \cref{sec:communication,sec:computation} offer supplementary information regarding the estimation of communication costs and computation costs, respectively. Further exploration into the efficacy of random feature approximation and the performance of our methods on the small-scale \cifar{} dataset are presented in \cref{sec:brrrf} and \cref{sec:cifar}, respectively. \cref{sec:best} provides additional plots that compare the best methods for \gld and \inat.

\subsection{\fedrrr Equivalence to (Centralized) \rr}
\label{sec:equivalence}
In this section, we evaluate the results of \fedrrr and \fedrrrrf on several splits of the \inat dataset, simulating various levels of statistical heterogeneity as proposed in ~\citet{hsu2020federated}. Both algorithms rely only on the pre-trained feature extractor to train the classifiers and do not adjust the representation. Hence, to ensure a fair comparison, we compare them with the \textit{Linear Probing} version (LP) of the FedAvg baseline, where we keep the parameters of the feature extractor frozen and train only the softmax classifier. For these experiments, we choose FedAvg as the baseline to compare with because it shows similar performances to FedAvgM and Scaffold fails to converge.

Specifically, \cref{fig:stat_het_inat} compares \fedrrr, \fedrrrrf with 10k random features, and FedAvg-LP, using four different \inat splits (for details on the splits, refer to \cref{tab:datasets} in \cref{sec:impl_det}). All the \fedrrr and \fedrrrrf 10k experiments converge to 45.1\% and 47.6\% accuracy, respectively, which are equivalent to \rr and \rr-RF with 10k random features in the centralized scenario (the dashed lines). This confirms the invariance to statistical heterogeneity and the equivalence of the FL and centralized solution, as discussed in \cref{sec:fed3rprops} from a theoretical perspective. Finally, as noticeable, convergence is much faster than FedAvg-LP, as its speed is proportional to the number of clients in the specific split, as discussed in \cref{sec:fed3rprops}.

\subsection{\fedrrr vs. Gradient-based FL Baselines}

\cref{fig:fed3r} shows how the methods perform in terms of \textit{accuracy} and \textit{convergence speed} (left), \textit{communication costs} to reach the target accuracy (center), and \textit{average computation} needed per client to reach the target accuracy (right). This is shown for both the \gld dataset (first row) and the \inat dataset (second row). The communication and computation costs can be interpreted as the budget needed for an FL system to reach a specific accuracy using the respective method.

As shown in \cref{fig:fed3r}, \fedrrr outperforms all the LP baselines in terms of speed and computational efficiency. 
Notably, on the \gld dataset, \fedrrr achieves comparable results to Scaffold-LP -- the best of the LP baselines -- while requiring two orders of magnitude less communication and computations, with \fedrrrrf even surpassing it at the expense of communication cost, but still being more computationally efficient. 

Remarkably, \fedrrr exhibits even greater efficacy on the \inat dataset, surpassing all LP and full-training baselines by a substantial margin across all evaluation criteria, including rounds, total communication, and computation costs. 
This underscores the significant impact of exact aggregation classifiers and the challenges faced by optimization-based FL methods in heterogeneous cross-device settings.

\paragraph{Discussion on the convergence speed.} Both \fedrrr and \fedrrrrf (with $D=5k$ and $D=10k$) are much faster than the baselines and require up to two orders of magnitude less communication and average computation budget. For instance, \fedrrrrf with $D=10k$ on the \gld dataset hits 40$\%$ accuracy after 27.3 rounds on average, compared to the 528.7 (speedup $\times$19.3) needed by FedAvg, 285.7 needed by Scaffold (speedup $\times$10.5), and 2251.3 (speedup $\times$82.4) and 690.33 (speedup $\times$25.3) needed by their corresponding LP versions. Similar consistent speedups can be observed for higher values of accuracy and on \inat.

\begin{figure}
    \centering
    \includegraphics[width=\linewidth]{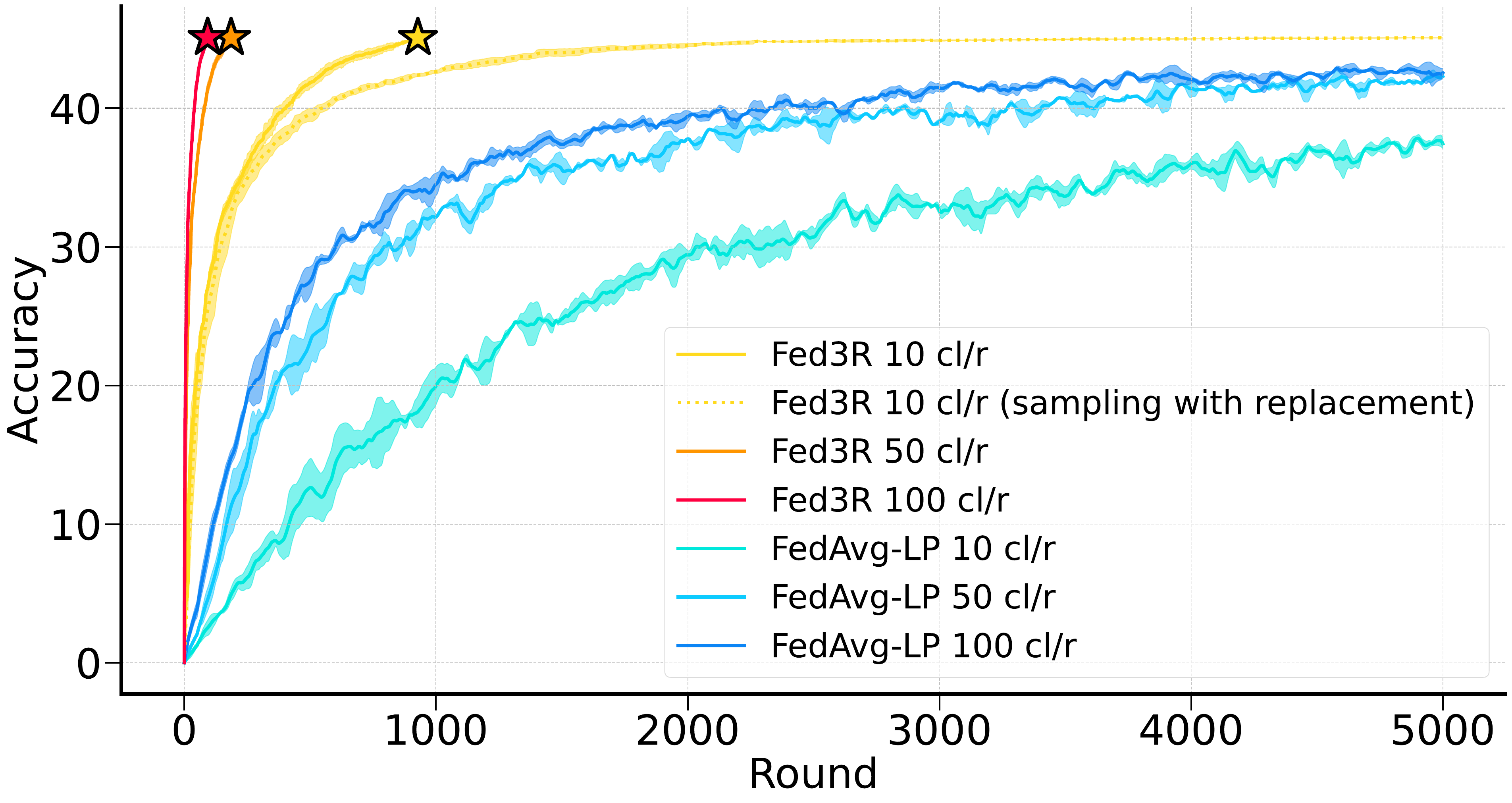}
    \caption{Accuracy vs Rounds with three different participation rates (indicated in the legend by $x$ cl/r, where cl/r stands for \textit{sampled clients per round}) and two sampling strategies (\textit{without replacement} for \fedrrr and \textit{with replacement} for FedAvg-LP, if not differently specified), \inat dataset. %
    }
    \label{fig:suboptimal}
\end{figure}

Indeed, since each client needs to be sampled only once, the convergence speed of \fedrrr depends only on the total number of clients and the participation rate. With 10 clients sampled per round and a total of 1262 clients for \gld and 9275 clients for \inat, \fedrrr always needs exactly 127 and 928 rounds to converge, respectively, which is an important advantage over gradient-based optimization methods as they require multiple passes over the data.

\paragraph{Performance with different sampling rates.} \cref{fig:suboptimal} shows how \fedrrr final performance is invariant to the number of clients sampled at each round by construction (as explained in \cref{sec:fed3rprops}). As a worst-case analysis, we also show that even sampling \textit{with replacement}, as in FedAvg and the other classical algorithms,  proves to be faster than the LP methods.
Notably, \fedrrr, with a sampling rate of 10 clients per round, converges faster than FedAvg-LP with a sampling rate of 100 clients per round. Indeed, \fedrrr almost achieves convergence performance after just 1.5k rounds. Therefore, \fedrrr does not really need to wait for all the clients in the federation to be available. For further investigation on how many rounds are needed to sample with replacement a given percentage of distinct clients, we refer the reader to the \cref{sec:coupons}.

\begin{table}[t]

    \caption{Final accuracy ($\%$) achieved by the \fedrrr family of classifiers and by the FedNCM classifier.}
    \label{tab:fedncm}
    \centering
    \begin{adjustbox}{width=\linewidth}
        \centering
        
        \begin{tabular}{lcccc}
            
            \toprule

            & \textbf{\fedrrr} & \textbf{\fedrrrrf 5k} & \textbf{\fedrrrrf 10k} & \textbf{\fedncm} \\
            
            \midrule
            
            \gld & 49.6 \tiny{$\pm$ 0.0} & 53.9 \tiny{$\pm$ 0.0} & \textbf{56.6 \tiny{$\pm$ 0.0}} & 36.2 \tiny{$\pm$ 0.0} \\
            \inat & 45.1 \tiny{$\pm$ 0.0} & 46.8 \tiny{$\pm$ 0.0} & \textbf{47.6 \tiny{$\pm$ 0.0}} & 32.2 \tiny{$\pm$ 0.0} \\
            
            \bottomrule
        
        \end{tabular}
    \end{adjustbox}
\end{table}

\paragraph{Ablation on Fed3R vs. FedNCM.} Similarly to \fedrrr, \citet{legate2023guiding} propose fitting a  closed-form classifier using Nearest Class Means (\fedncm). \cref{tab:fedncm} compares the performance of \fedncm, \fedrrr, and \fedrrrrf at convergence for both the \gld and \inat datasets, without the fine-tuning stage. 
\fedrrr clearly emerges as a more powerful and robust approach that can deal with complex datasets, outperforming FedNCM by a significant margin - up to 20 accuracy points with the kernelized version. 
Consequently, as our method yields superior classifiers, we omit the \fedncm experiments from the FT discussions in \cref{sec:fed3rft-exp}. This decision is based on the assumption that employing a weaker classifier as initialization would result in a lower final accuracy.

\subsection{\fedrrrft experiments}
\label{sec:fed3rft-exp}

\begin{figure*}
    \centering
    \begin{subfigure}[b]{0.33\linewidth}
        \includegraphics[width=\linewidth]{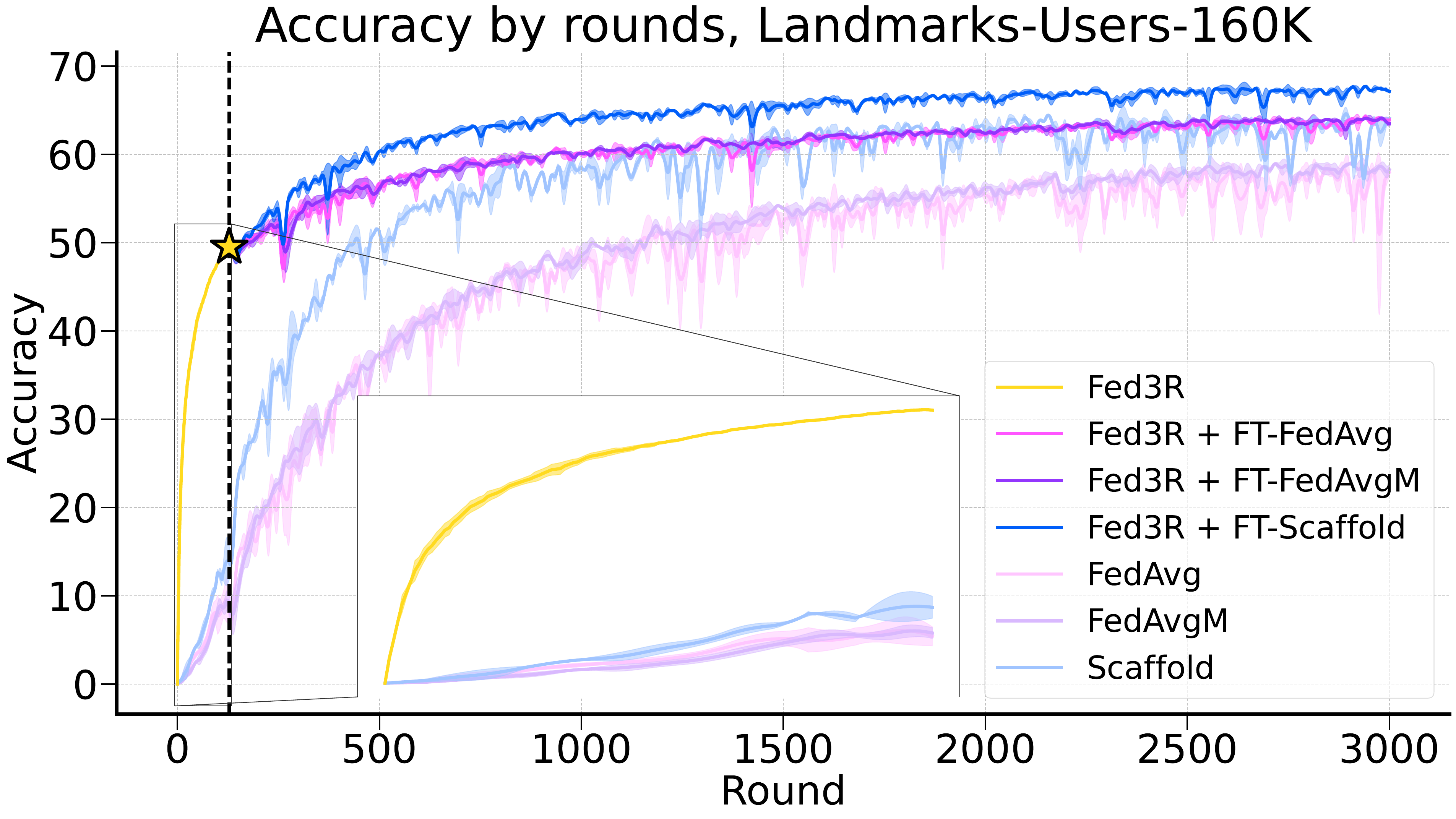}
    \end{subfigure}
    \begin{subfigure}[b]{0.33\linewidth}
        \includegraphics[width=\linewidth]{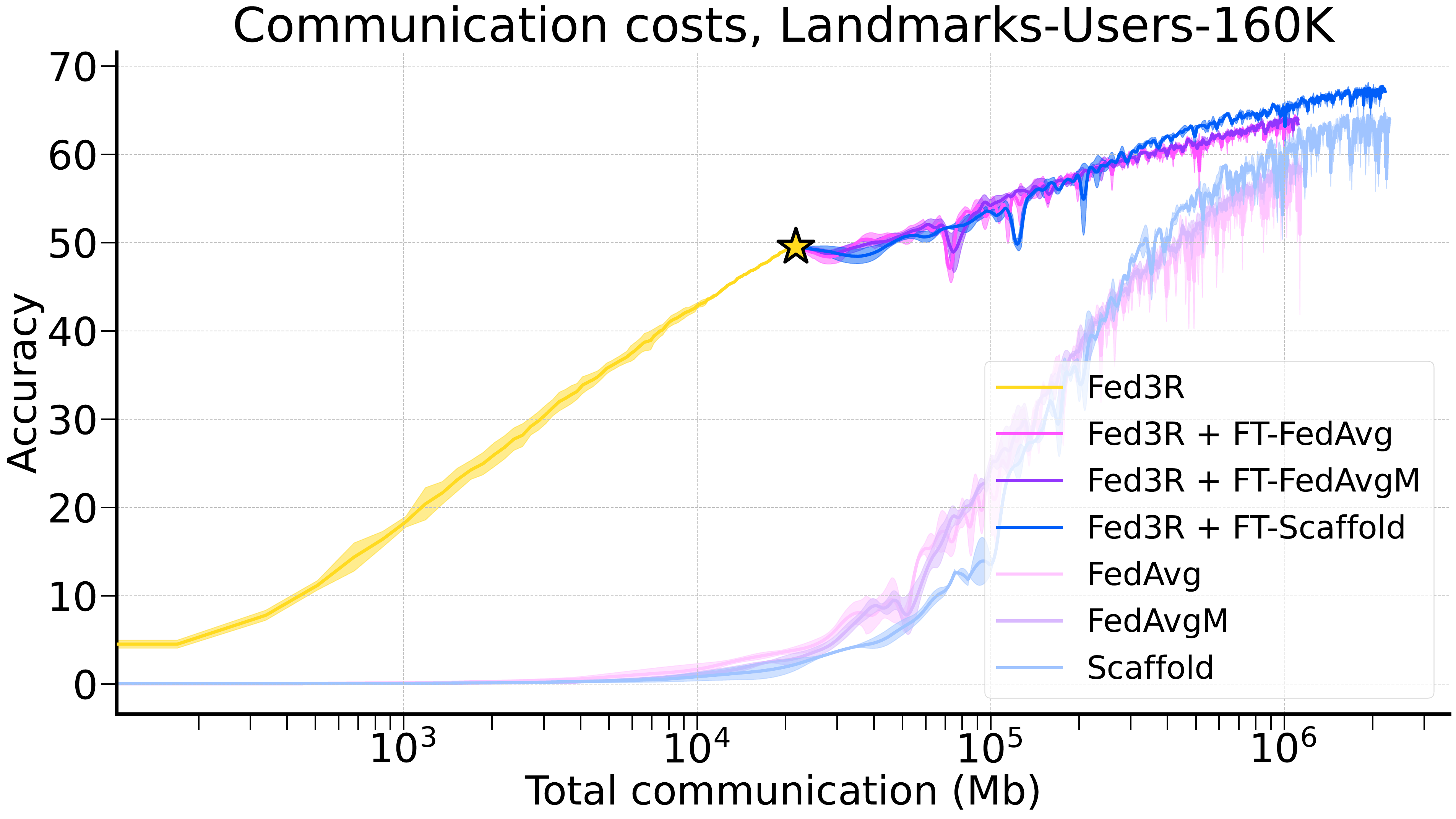}
    \end{subfigure}
    \begin{subfigure}[b]{0.33\linewidth}
        \includegraphics[width=\linewidth]{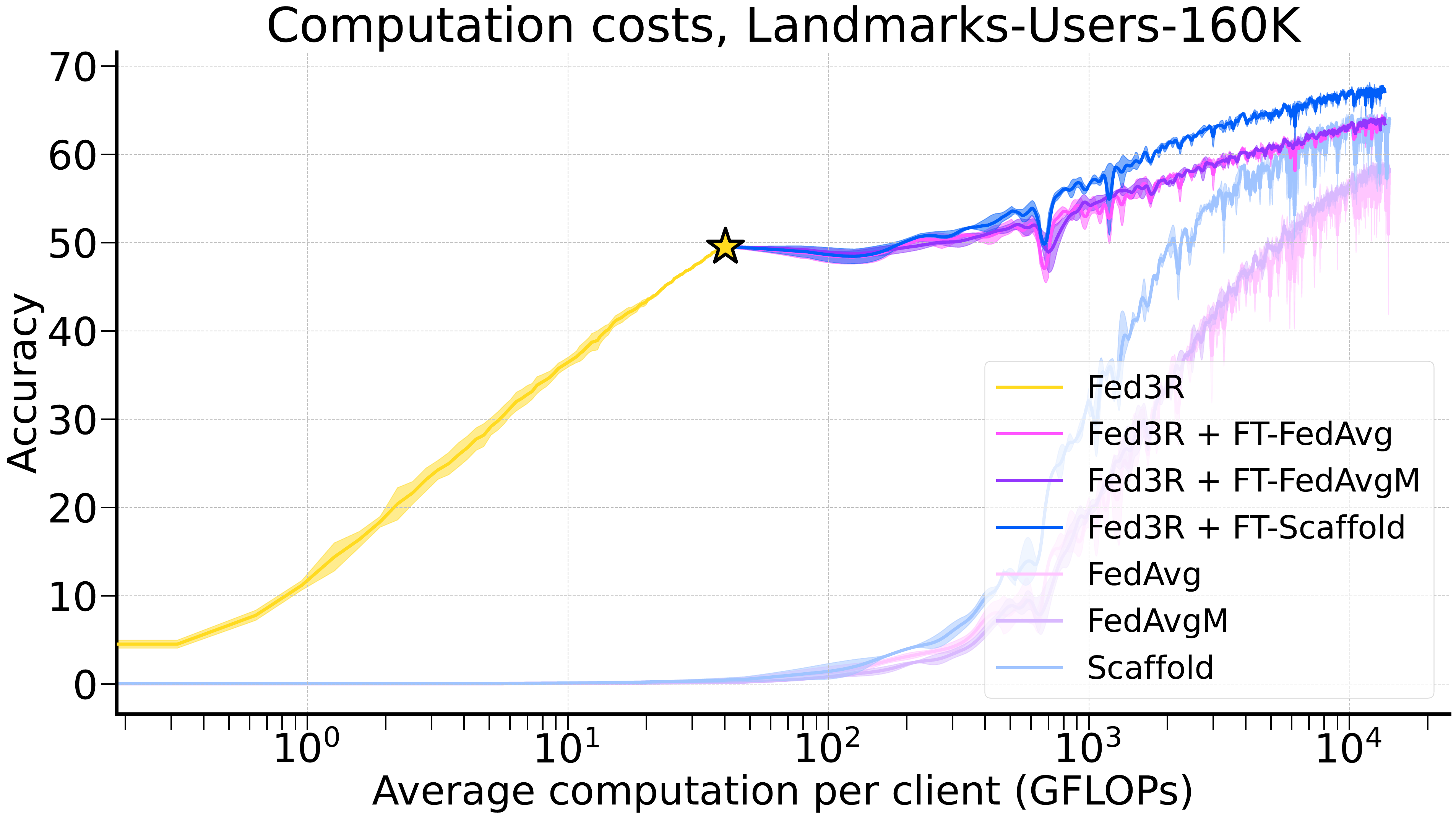}
    \end{subfigure}
    \caption{Comparison between \fedrrrft and the baselines \gld dataset. At the convergence point of \fedrrr, we substitute the parameters of the \fedrrr classifier to the ones of the softmax and then use another algorithm for fine-tuning.}
    \label{fig:fed3rftf}
\end{figure*}

\begin{figure*}
    \centering
    \begin{subfigure}[b]{0.33\linewidth}
        \includegraphics[width=\linewidth]{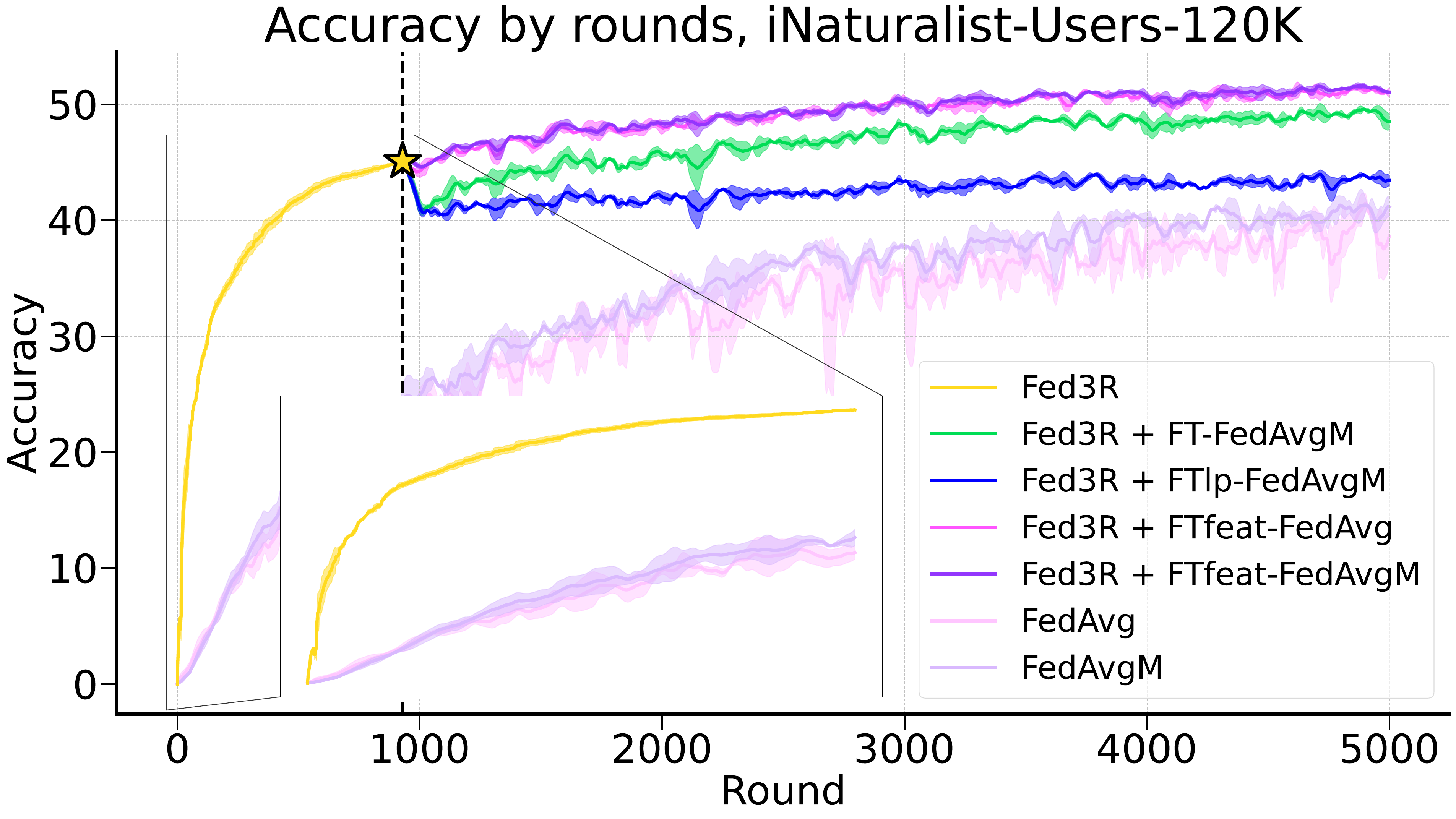}
    \end{subfigure}
    \begin{subfigure}[b]{0.33\linewidth}
        \includegraphics[width=\linewidth]{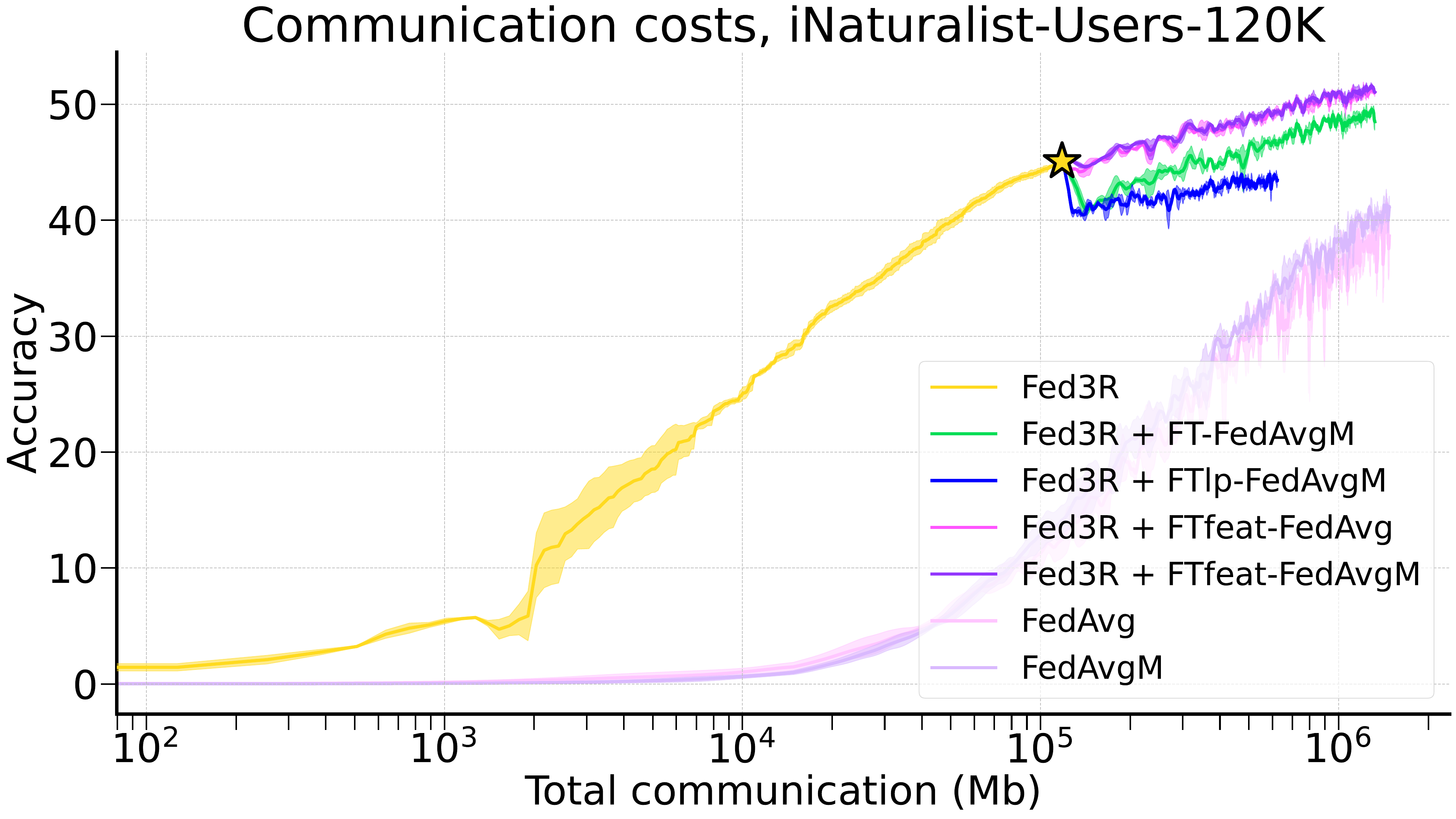}
    \end{subfigure}
    \begin{subfigure}[b]{0.33\linewidth}
        \includegraphics[width=\linewidth]{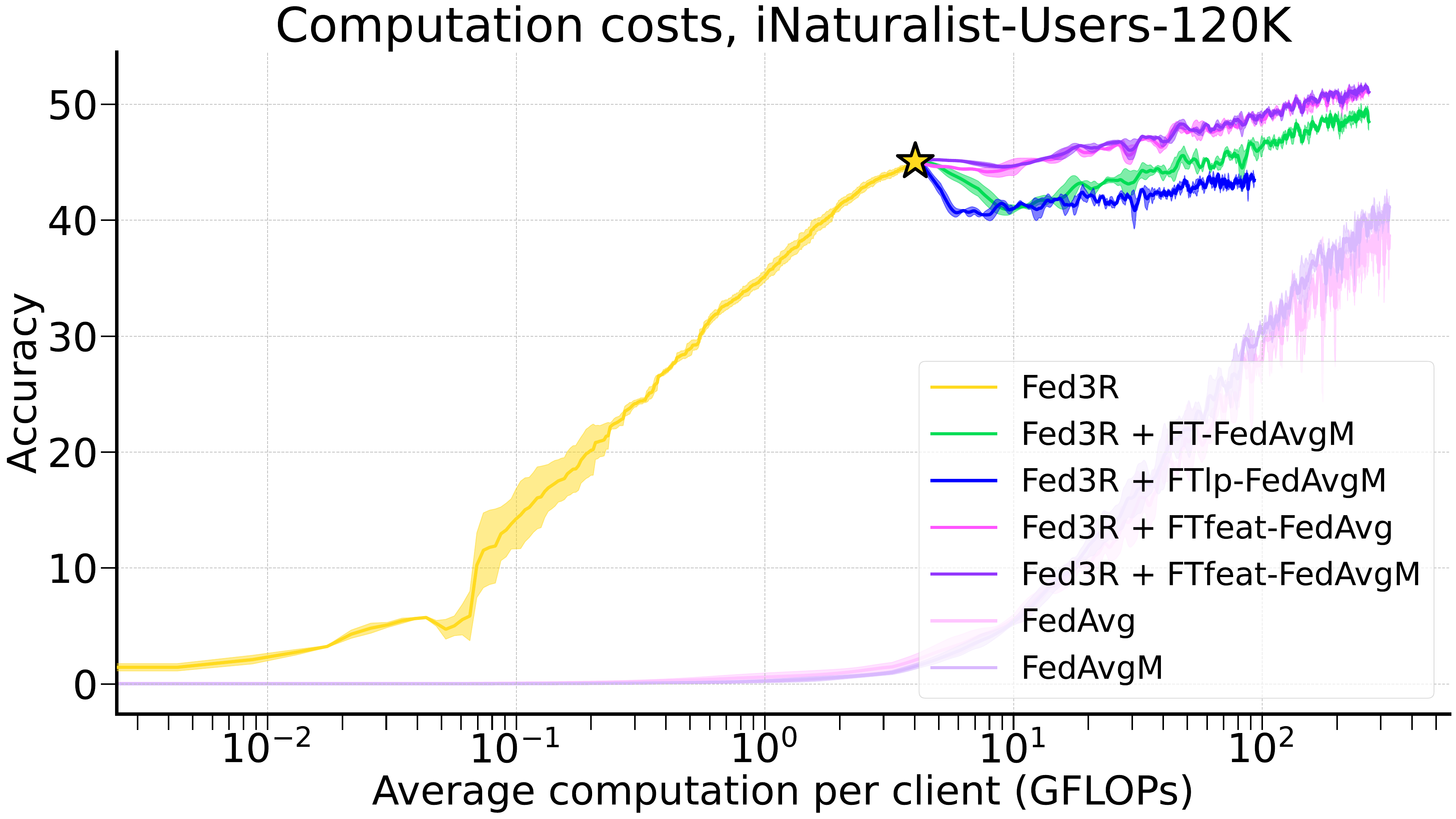}
    \end{subfigure}
    \caption{Comparison between \fedrrrft in all its variants and the baselines, \inat dataset.}
    \label{fig:fed3rft}
\end{figure*}

In the previous section, we showed the efficacy of the \fedrrr algorithm in terms of speed and efficiency. However, sometimes its performance may not surpass baseline methods, particularly when utilizing \fedrrr instead of \fedrrrrf, as shown in \cref{fig:fed3r}, top row. However, while employing \fedrrrrf incurs a minimal computational overhead, it substantially escalates communication costs, rivaling other baseline methods. Moreover, both \fedrrr and \fedrrrrf rely only on pre-trained features, as the pre-trained feature extractor is not optimized on the target datasets of the clients. Therefore, we propose also \fedrrrft and its variants, as discussed in \cref{sec:fed3rft}.

\begin{table}[t]
    
    \caption{\fedrrrft final performance (Acc. $\%$).}
    \label{tab:cmp-ft}
    \centering
    \begin{adjustbox}{width=\linewidth}
        \centering
    
        \begin{tabular}{cccccc}
        
            \toprule
            
            \textbf{Dataset} & \makecell{\textbf{FT alg.}} & \makecell{\textbf{Classifier} \\ \textbf{Initialization}} & \textbf{\textsc{FTfeat}} & \textbf{\textsc{FTlp}} & \textbf{\textsc{FT}} \\
            
            \midrule

            \multirow{6}{*}{\rotatebox[origin=c]{90}{\makecell{Landmark- \\ Users-160K} \hspace{.8em}}} & \multirow{2}{*}{FedAvg} & \xmark & - & 41.0 \tiny{$\pm$ 1.6} & 57.7 \tiny{$\pm$ 1.2} \\
             & & \fedrrr & 59.6 \tiny{$\pm$ 0.2} & 56.7 \tiny{$\pm$ 0.4} & \textbf{64.1 \tiny{$\pm$ 0.4}} \\
            
            \cmidrule{2-6}
            
             & \multirow{2}{*}{FedAvgM} & \xmark & - & 40.8 \tiny{$\pm$ 1.2} & 58.7 \tiny{$\pm$ 0.8} \\
             & & \fedrrr & 59.0 \tiny{$\pm$ 0.2} & 56.2 \tiny{$\pm$ 0.5} & \textbf{64.1 \tiny{$\pm$ 0.2}} \\
            
            \cmidrule{2-6}
            
             & \multirow{2}{*}{Scaffold} & \xmark & - & 51.7 \tiny{$\pm$ 0.3} & 63.4 \tiny{$\pm$ 0.9} \\
             & & \fedrrr & 63.4 \tiny{$\pm$ 0.1} & 58.0 \tiny{$\pm$ 1.5} & \textbf{67.4 \tiny{$\pm$ 0.3}} \\

            \midrule

            \multirow{4}{*}{\rotatebox[origin=c]{90}{\makecell{iNaturalist- \\ Users-120K} \hspace{.01em}}} & \multirow{2}{*}{FedAvg} & \xmark & - & 36.7 \tiny{$\pm$ 0.4} & 39.5 \tiny{$\pm$ 3.2} \\
             & & \fedrrr & \textbf{50.8 \tiny{$\pm$ 0.2}} & 42.0 \tiny{$\pm$ 0.3} & 49.0 \tiny{$\pm$ 0.6} \\
            
            \cmidrule{2-6}
            
             & \multirow{2}{*}{FedAvgM} & \xmark & - & 37.6 \tiny{$\pm$ 0.2} & 39.3 \tiny{$\pm$ 0.7} \\
             & & \fedrrr & \textbf{51.5 \tiny{$\pm$ 0.2}} & 43.5 \tiny{$\pm$ 1.9} & 49.8 \tiny{$\pm$ 0.8} \\
            
            \bottomrule
    
        \end{tabular}
    \end{adjustbox}

\end{table}

Table \ref{tab:cmp-ft} shows the final accuracy values for the different strategies. Moreover, \cref{fig:fed3rftf} shows the performance and the costs of \fedrrrft and baseline methods for the \gld dataset, while \cref{fig:fed3rft} compares the three variants of \fedrrrft for the \inat dataset. FedAvgM serves as the FT algorithm in both scenarios. Notably, at least one of our \fedrrrft variants significantly outperforms the baselines for both datasets, achieving accuracies of 67.4 $\pm$ 0.3 and 51.5 $\pm$ 0.2 in the \gld and \inat experiments, respectively, and the curves associated with our methods are consistently and significantly above the comparisons across every x-axis value. 

Fine-tuning the entire model shows benefits on \gld, which is more similar to cross-silo FL than \inat. On the other hand, in federated settings with more clients, such as in \inat, there is a significant negative impact during the aggregation phase for the \fedrrrft and \fedrrrftc experiments, as the classifier is fine-tuned and becomes susceptible to the classifier bias phenomenon, as discussed in Section \ref{sec:intro}. Conversely, keeping the classifier fixed and only fine-tuning the feature extractor as in the \fedrrrftf experiments eliminates this phenomenon for the classifier, ensuring performance improvement and clearly indicating that the pre-trained features were not sufficiently good for the target task.

\subsection{Features Quality Evaluation via Ridge Regression}

In this section, we show that \rr is a useful tool for quantifying features' quality and linear separability in an FL scenario. Indeed, \rr provides a closed-form deterministic solution that solely depends on the feature space and that can be computed in federated settings through the \fedrrr equivalent formulation, unaffected by statistical heterogeneity. Moreover, \rr is independent of the specific training hyper-parameters, conversely to softmax classifiers trained with gradient-based methods.

\begin{table}[t]

    \caption{Quality of the feature extractors (acc. $\%$) at convergence measured with Ridge Regression.
    }
    \label{tab:fextr}
    \centering
    \begin{adjustbox}{width=\linewidth}
        \centering
        
        \begin{tabular}{cccccc}
            
            \toprule

               \textbf{Dataset} & \textbf{FT alg.} & \textbf{FT strategy} & \textbf{Cls Init.} & \textbf{Softmax} & \textbf{\rr}  \\
            
            \midrule

                \multirow{7}{*}{\rotatebox[origin=c]{90}{\makecell{Landmarks- \\ Users-160K} \hspace{.5em}}} & - & - & \fedrrr & - & 49.6 \tiny{$\pm$ 0.0} \\

                \cmidrule{2-6}
                
                & \multirow{3}{*}{FedAvg} & \textsc{feat + lp} & - & 57.7 \tiny{$\pm$ 1.2} & 58.8 \tiny{$\pm$ 2.1} \\
                & & \textsc{feat + lp} & \fedrrr & \textbf{64.1 \tiny{$\pm$ 0.4}} & 59.6 \tiny{$\pm$ 0.2} \\
                & & \textsc{feat} & \fedrrr & 59.6 \tiny{$\pm$ 0.2} & \textbf{62.1 \tiny{$\pm$ 0.1}} \\

                \cmidrule{2-6}
                
                & \multirow{3}{*}{Scaffold} & \textsc{feat + lp} & - & 63.4 \tiny{$\pm$ 0.9} & 57.0 \tiny{$\pm$ 1.9} \\
                & & \textsc{feat + lp} & \fedrrr & \textbf{67.4 \tiny{$\pm$ 0.3}} & 61.8 \tiny{$\pm$ 0.1} \\
                & & \textsc{feat} & \fedrrr & 63.4 \tiny{$\pm$ 0.1} & \textbf{64.3 \tiny{$\pm$ 0.2}} \\
                
            \midrule %

                \multirow{4}{*}{\rotatebox[origin=c]{90}{\makecell{iNaturalist- \\ Users-120K} \hspace{.05em}}} & - & - & \fedrrr & - & 45.1  \tiny{$\pm$ 0.0} \\

                \cmidrule{2-6}
                
                & \multirow{3}{*}{FedAvg} & \textsc{feat + lp} & - & 39.5 \tiny{$\pm$ 3.2} & 53.1 \tiny{$\pm$ 0.9} \\
                & & \textsc{feat + lp} & \fedrrr & 49.0 \tiny{$\pm$ 0.6} & 52.2 \tiny{$\pm$ 0.3} \\
                & & \textsc{feat} & \fedrrr & \textbf{50.8 \tiny{$\pm$ 0.2}} & \textbf{54.6 \tiny{$\pm$ 0.1}} \\
                
            \bottomrule
        
        \end{tabular}
    \end{adjustbox}
\end{table}

\cref{tab:fextr} provides this quantitative analysis, where we compute \rr on the feature extractor of the model after fine-tuning, at convergence. The findings demonstrate an enhancement in the quality of the learned features with the robust \fedrrr initialization. This initialization aids in stabilizing the training process, reducing destructive interference and forgetting caused by heterogeneity. %
Specifically, \fedrrrft and \fedrrrftf consistently yield higher \rr accuracy than the corresponding baseline with the same fine-tuning algorithm, for both the \gld and \inat datasets. Moreover, \fedrrrftf consistently outperforms \fedrrrft in all cases, as keeping the classifier fixed completely prevents the classifier bias.

Furthermore, in realistic cross-device scenario as \inat, \rr at convergence achieves even higher accuracy than the softmax classifier in all the fine-tuning strategies. This observation suggests the possibility of executing \fedrrr after the training process to further improve performance.

\section{Conclusion}

In this work, we introduce \fedrrr, a family of Federated Learning algorithms based on Recursive Ridge Regression. \fedrrr is designed to minimize communication and computation costs and accelerate convergence speed while adhering to the privacy constraints of FL. Unlike gradient-based FL algorithms where statistical heterogeneity is a significant challenge, \fedrrr is immune to statistical heterogeneity by design and can also serve as a robust initialization for further fine-tuning with optimization-based FL algorithms.
Results show that our algorithm requires up to two orders of magnitude less communication and computation costs to convergence than the baselines (see \cref{fig:fed3r}) and improves the accuracy up to 12\% in challenging cross-device FL scenarios (see \cref{tab:cmp-ft}, \inat results).
Finally, our findings reveal that the features produced during the fine-tuning stage are more robust than those achieved by other methods at convergence {see \cref{tab:fextr}). This underscores the notion that in challenging cross-device settings, the quality of the feature extractor may serve as a bottleneck alongside the classifier's quality. 
Future works may extend \fedrrr to streaming data or personalized learning scenarios within the FL framework.

\section*{Impact Statement}

\fedrrr significantly enhances training efficiency by providing remarkable speed and minimal computational and communication costs. By lightening the FL training load, our algorithm not only improves efficiency but also reduces the energy required for training. This not only benefits cost savings but also contributes to reducing environmental pollution associated with training models, as the energy required for training may still be generated using unsustainable methods. This has the potential to significantly impact various applications across industries, making them more accessible and cost-effective. The rapid execution and resource efficiency of our method could lead to increased adoption of FL techniques, enabling advancements in fields ranging from healthcare and finance to autonomous systems and beyond.

\section*{Acknowledgments}

This study was carried out within the FAIR - Future Artificial Intelligence Research and received funding from the European Union Next-GenerationEU (PIANO NAZIONALE DI RIPRESA E RESILIENZA (PNRR) – MISSIONE 4 COMPONENTE 2, INVESTIMENTO 1.3 – D.D. 1555 11/10/2022, PE00000013). This manuscript reflects only the authors’ views and opinions, neither the European Union nor the European Commission can be considered responsible for them.
We acknowledge the CINECA award under the ISCRA  initiative for the availability of high-performance computing resources and support. We also thank the reviewers and area chair for their valuable comments.

\bibliography{IEEEabrv, bibfile}
\bibliographystyle{icml2024}

\newpage
\appendix
\onecolumn

\section{Additional Related Works}
\label{sec:more_related}

In this section, we expand on the related works concerning Ridge Regression in Distributed and Federated Learning settings and Transfer Learning methods involving pre-trained models in FL.

\paragraph{Ridge Regression in distributed and federated learning.} 
Regarding prior works on Ridge Regression in FL settings, \citet{afonin2021towards} finds an optimal Tikhonov Regularized Least Squares solution for a federation of only two clients that, in practice, constitutes a cross-knowledge distillation framework. 
\citet{cai2022efficient} and \citet{huang2022coresets} apply Ridge Regression to a \textit{Vertical FL} scenario in which 
each client possesses different features of all the samples. Conversely, in this work, we focus on the more common \textit{Horizontal} FL~\cite{yang2019federated} setting, where the feature space is shared among clients, but local datasets vary. The most critical distinction between Federated \rr methods for V-FL and H-FL concerns how best to compute gradients and aggregate statistics in a privacy-preserving manner while having the global dataset partitioned in a fundamentally different way across clients. Such radically different splitting strategies result in distinct algorithm design choices. For example, client drift is a major challenge in H-FL, and algorithms aim to reduce the effect of biased local gradients during aggregation. On the other hand, V-FL methods mostly focus on reducing the communication costs for computing good loss and gradient estimates based on a reduced feature set while preserving privacy. Importantly, we also have to face severe statistical heterogeneity in H-FL, which is typically not an issue in V-FL since it concerns only Cross-Silo scenarios. To the best of our knowledge, we are the first to apply \rr within this specific context by leveraging its online formulation as an alternative to gradient-based optimization to speed up training and improve communication efficiency in realistic heterogeneous cross-device scenarios. 

Other works tackle large-scale least-squares problems in distributed learning and optimization settings \cite{zhang2012communication, DeCKRR, verticalDL}. While similar in spirit, distributed settings differ fundamentally from FL as privacy is not a constraint, and data is usually assumed i.i.d. across clients. 

\paragraph{Transfer learning methods with pre-trained models in FL.} 

The work presented in \citet{oh2021fedbabu} proposes a two-stage algorithm that uses a fixed, random classifier and trains only the feature extractor. However, while \citet{oh2021fedbabu} focuses on the Personalized FL setting by specializing in the head for each client, our work addresses the classifier bias problem in the conventional FL setting, where the goal is to learn a global classifier that represents the overall underlying distribution. Similarly, \citet{kim2024navigating} outlines a method for semi-supervised learning scenarios and object detection to selectively train the model's backbone while keeping the rest of the model frozen. They claim this approach helps train more consistent representations and establishes a stronger backbone for further fine-tuning with an extra regularization term.

Transfer learning techniques have recently garnered attention in the FL community to make Foundation Models suitable for the cross-device setting \cite{guo2023promptfl, chen2022fedtune, zhang2024towards}. These techniques exploit methods based on Low-Rank Approximation for parameter-efficient fine-tuning \cite{babakniya2023slora, cho2024heterogeneous, yi2023fedlora}.

\section{Privacy of \fedrrr}
\label{sec:privacy}

In the context of \fedrrr, clients have to transmit only the $A_k$ and $b_k$ statistics. Some may express concerns about the potential information leakage inherent in sharing this data, which extends beyond the disclosure associated with merely sharing model weights or gradients. However, it is crucial to note that any information the clients send to the server only needs to be aggregated. In other words, the server does not necessitate accessing individual values but rather needs solely to use the aggregated results. Therefore, privacy can be easily achieved by employing the Secure Aggregation protocol \cite{bonawitz2016practical}.

\section{Additional Implementation Details}
\label{sec:impl_det}

\fedrrr requires only one communication round with each client, which can occur as soon as the clients are ready. However, to guarantee privacy, we simulate the server waiting for a group of clients, similar to classical algorithms. In this way, a practical implementation might incorporate a Secure Aggregation \cite{bonawitz2016practical} step, where the information provided by individual clients is concealed within the aggregation of statistics shared by all sampled clients.

We run all the experiments using an NVIDIA A100-SXM4-40GB using the FL clients partitions provided by \cite{hsu2020federated} for \gld \cite{weyand2020google} and \inat \cite{van2018inaturalist}. For the cifar100 \cite{krizhevsky2009learning} experiments, we focus on the most heterogeneous case ($\alpha = 0$, \cite{hsu2019measuring}), where each client has access to images belonging to the same single class.

\begin{table}[t]

    \caption{Datasets additional information.}
    \label{tab:datasets}
    \centering
    \begin{adjustbox}{width=.5\linewidth}
        \centering
        
        \begin{tabular}{lccc}
            
            \toprule

            \textbf{Dataset} & \makecell{\textbf{Avg. samples} \\ \textbf{per client}} & $K$ & $C$ \\
            
            \midrule

            \landmark & 119.9 & 1262 & 2028 \\
            \inaturalist & 13.0 & 9275 & 1203 \\
            iNaturalist-Geo-100 & 33.4 & 3606 & 1203 \\
            iNaturalist-Geo-300 & 99.6 & 1208 & 1203 \\
            iNaturalist-Geo-1K & 326.9 & 368 & 1203 \\
            \cifar{} & 500 & 100 & 100 \\
            
            \bottomrule
        
        \end{tabular}
    \end{adjustbox}
    
\end{table}

Details on the datasets are provided in \cref{tab:datasets}. We used a MobileNetV2 \cite{mobilenetv2} network pre-trained on ImageNet for all our experiments. We replicate the same augmentation used to pre-train the model to best exploit the pre-trained features. Therefore, we scaled all the images to 224$\times$224, even for the 32$\times$32 images of \cifar{}.

We conducted the \gld experiments for 3000 rounds, the \inat experiments for 5000 rounds, and the \cifar{} experiments for 1500 rounds. We sampled 10 clients per round in all three cases unless stated otherwise. We utilized SGD as the client optimizer with a learning rate (lr) of 0.1 and a weight decay (wd) of $4 \times 10^{-5}$, a batch size of 50, and 5 local epochs for both \gld and \inat, and 1 local epoch for \cifar{}. Additionally, we employed SGD as the server optimizer \cite{reddi2020adaptive} with a learning rate (slr) set to 1.0 and no momentum (smom). The best hyper-parameters were the same across all methods and datasets, selected based on a grid search: $lr=\{0.1, 0.01\} \times slr=\{0.1, 1.0\} \times wd=\{0.0, 4 \cdot 10^{-5}\} \times smom=\{0.0, 0.9\}$. For the \fedrrr $\lambda$ hyper-parameter, we set $\lambda = 0.01$ as it consistently yielded the best \rr results. 

Despite our best efforts and hyper-parameters tuning, Scaffold failed to converge on the \inat dataset, although it converges on the \gld experiments. We attribute this to its initial design for cross-silo settings, as control variates become stale in realistic cross-device scenarios~\cite{karimireddy2020mime}.

\begin{figure*}
    \centering
    \begin{subfigure}[b]{0.49\linewidth}
        \includegraphics[width=\linewidth]{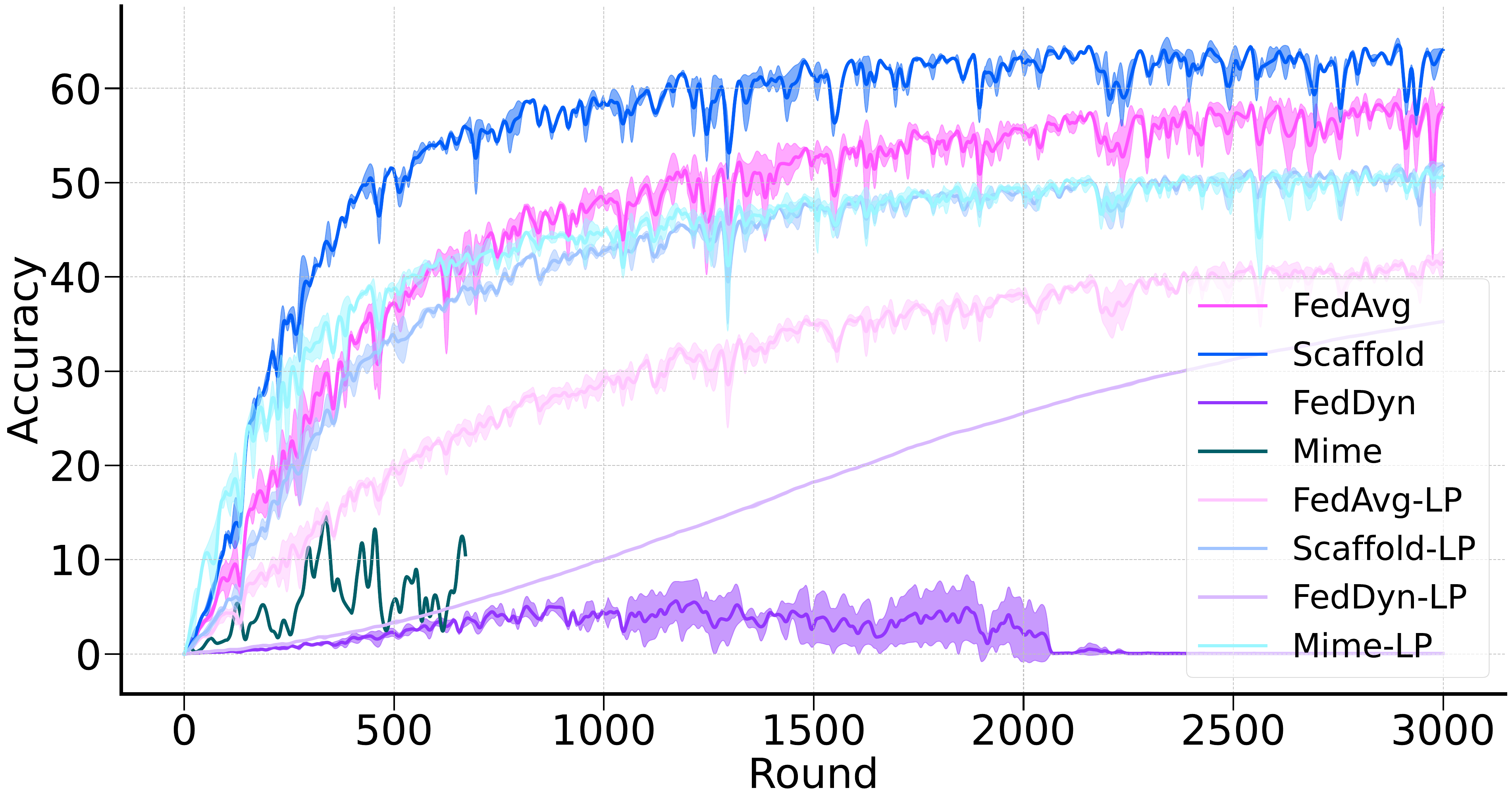}
    \end{subfigure}
    \begin{subfigure}[b]{0.49\linewidth}
        \includegraphics[width=\linewidth]{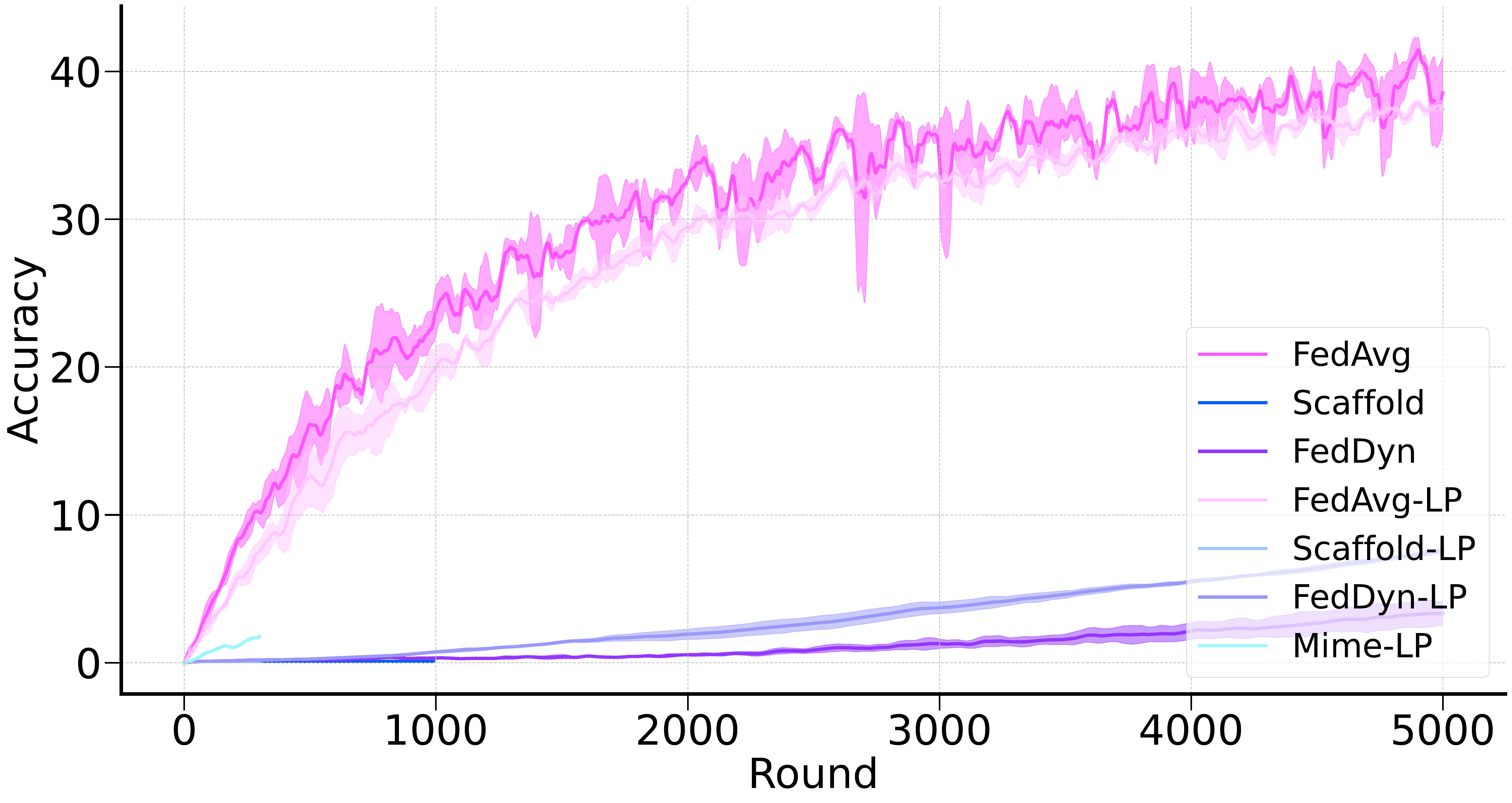}
    \end{subfigure}
    \caption{Comparison between baseline algorithms and their LP versions. Left: \landmark; right: \inaturalist.}
    \label{fig:baselines}
\end{figure*}

\cref{fig:baselines} shows the performance of the baseline algorithms with the best hyper-parameters from the grid search. Only FedAvg (and FedAvgM, which had similar results to FedAvg and was not included in the two plots for clarity) consistently performs well on both datasets. At the same time, the other algorithms struggle, especially in the realistic cross-device scenario of \inat, where FedAvg is the only baseline algorithm that works.

In all our \fedrrrrf experiments, we considered a random features approximation of an RBF kernel $k(z, \zeta) = e^{-\nicefrac{\norm{z - \zeta}^2}{2\sigma^2}}, \text{ } z, \zeta \in \R^d$. The hyper-parameter $\sigma$ has been tuned once in the centralized \gld setting, and the best value $\sigma=1000$ has been selected for all the experiments. For the \fedrrrft, \fedrrrftc, \fedrrrftf experiments, we found that the softmax temperature value of 0.1 yields the best results on both the \gld and \inat datasets, as \cref{fig:temp} shows.

\begin{figure}
    \centering
    \caption{Cross-entropy loss values evaluated on the training set using different softmax temperatures. The model is initialized with the \fedrrr classifier and the pre-trained feature extractor. The best temperature is 0.1 for both the \gld and \inat datasets.}
    \includegraphics[width=.4\linewidth]{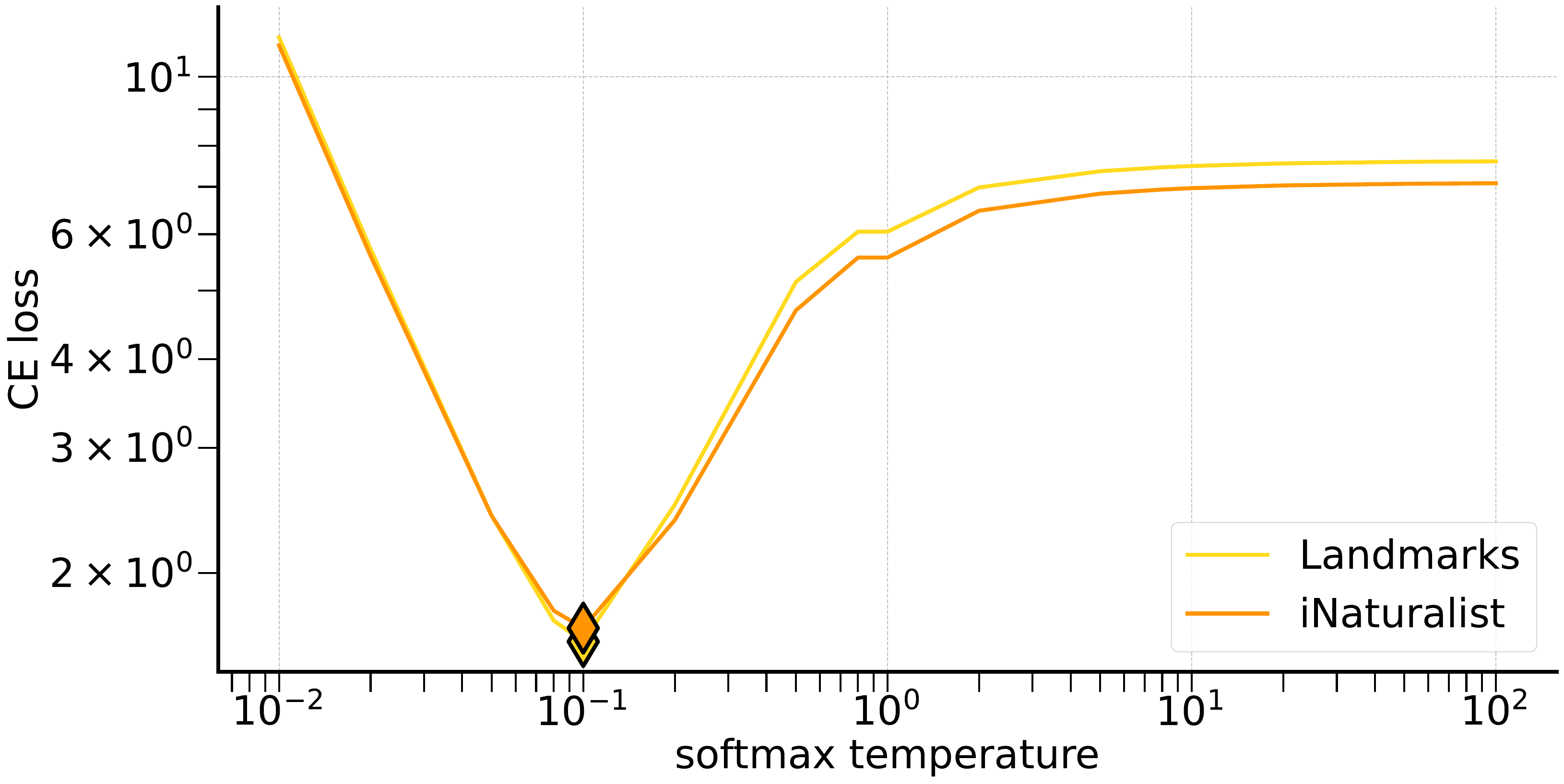}
    \label{fig:temp}
\end{figure}

We assume all the values are stored as FP32 numbers to estimate the communication and computation costs. See \cref{sec:communication} and \cref{sec:computation} for more details.

\section{Communication Costs Computation}
\label{sec:communication}

We initially estimate the costs per round for each client to evaluate the communication costs. The overall communication cost per round and client comprises two components: the downstream and upstream costs, representing the communication from and to the server, respectively, which may vary across different methods. With this figure in hand, we then determine the total costs per round by multiplying the cost per round per client by the number of sampled clients per round. In all the communication costs plots, we multiplied the final values by 4 to measure the final cost in bytes, as we assume all the parameters are stored as FP32 values, \ie 4 bytes.

Below, we briefly summarize how the downstream and upstream costs have been calculated per each algorithm and eventual additional costs. Let $m$, $b$, and $c$ be the sizes of the whole model, the feature extractor, and the classifier, respectively. As the classifier is a linear layer, its size is equivalent to the product of the latent feature dimensionality and the number of classes of the dataset: $c = dC$. Therefore, $m = b + dC$. Then:
\begin{itemize}[leftmargin=*]
    \item \textbf{FedAvg \cite{mcmahan2017communication}, FedAvgM \cite{hsu2019measuring}.} Each sampled client downloads and uploads the model only once: \textit{Downstream/k} = $b + dC$, \textit{Upstream/k} = $b + dC$.
    \item \textbf{Scaffold \cite{karimireddy2020Scaffold}.} Each sampled client downloads and uploads both the model and its control variate: \textit{Downstream/k} = $2(b + dC)$, \textit{Upstream/k} = $2(b + dC)$.
    \item \textbf{FedAvg-LP, FedAvgM-LP.} Each sampled client downloads and uploads the classifier only once: \textit{Downstream/k} = $dC$, \textit{Upstream/k} = $dC$.
    \item \textbf{Scaffold-LP.} Each sampled client downloads and uploads both the classifier and its control variate: \textit{Downstream/k} = $2dC$, \textit{Upstream/k} = $2dC$.
    \item \textbf{\fedrrr, \fedrrrrf.}
    Each client needs to receive the feature extractor parameters only once. If we do not assume clients already have the feature extractor parameters before the training begins (though this assumption is reasonable in scenarios where the server, as a business, deploys its application and may have already incorporated these parameters in the clients' software), there is an additional communication cost of $bK$. Except that for these costs, each sampled client does not need to download any information from the server, but it needs to upload the local statistics $A_k, b_k$ to the server: \textit{Downstream/k} = $0$, \textit{Upstream/k} = $d^2 + dC$. If we use \fedrrrrf the upstream costs per client are \textit{Upstream/k} = $D^2 + DC$ instead.
    \item \textbf{\fedrrrft}. In this scenario, the \textit{Downstream/k} and \textit{Upstream/k} costs correspond to those of \fedrrr during the initial phase of the experiments, when \fedrrrft generates the \fedrrr classifier as initialization for the softmax classifier. Subsequently, the costs reflect those of the FT algorithm, with the sole exception for \fedrrrftf, where the FT phase costs are \textit{Downstream/k} = \textit{Upstream/k} $= b$ for FedAvg and FedAvgM, and \textit{Downstream/k} = \textit{Upstream/k} $= 2b$ for Scaffold.
\end{itemize}

\section{Average Computation Costs per Client}
\label{sec:computation}

We prioritize the average computation costs per client over the total computation cost among all clients. This decision stems from our belief that this statistic provides more insightful information regarding the budget required by an FL system developer for their clients.

To estimate this value, let $\mathcal{T}$ be the total average cost \emph{per round} per single client for a given algorithm. Let $\boldsymbol{x}$ be the number of times a specific client is sampled.
Then, the cumulative average cost $\mathcal{T}_t$ from round $1$ to round $t$ is proportional to the expected number of times $\E[\boldsymbol{x}]$ a specific client is sampled over $t$ rounds, that is $\E[\boldsymbol{x}] = t \frac{\kappa}{K}$, where $\kappa$ is the number of clients sampled per each round and $K$ is the total number of clients. Therefore, $\mathcal{T}_t = \mathcal{T}\E[\boldsymbol{x}] = \mathcal{T}t\frac{\kappa}{K}$.

The specific $\mathcal{T}$ value depends on the algorithm. Let $F_*$ and $B_*$ be the costs of one forward pass of a single image and one backward pass of a single image through the model $*$. As the authors of \cite{legate2023guiding}, we approximate $B_* \simeq 2F_*$, and consider both the forward $F^N_*$ and backward $B^N_*$ of a batch of $N$ images as directly proportional to $B$ and $F$, \ie $F^N_* = NF_*$ and $B^N_* = NB_*$. Therefore, one epoch's total forward and backward costs for a single client are simply $F^{n_k}_*$ and $B^{n_k}_*$.

We measure these costs in FLOPs. Therefore, we divide by half the count of the matrix operations since one FLOP is defined as one addition and one multiplication of floating point numbers. Let $E$ be the number of local epochs. Then:

\begin{itemize}[leftmargin=*]
    \item \textbf{FedAvg \cite{mcmahan2017communication}, FedAvgM \cite{hsu2019measuring}, Scaffold \cite{karimireddy2020Scaffold}.} All these methods have one forward pass and one backward pass through the whole model and other negligible operations, such as SGD updates and computations of client control variates for Scaffold. Therefore, we consider the same total cost per round $\mathcal{T} = En_k(F_\mathcal{M} + B_\mathcal{M}) = 3En_kF_\mathcal{M}$.
    \item \textbf{FedAvg-LP, FedAvgM-LP, Scaffold-LP.} In this case, the forward is through the whole model, but the backward is only up to the classifier: $\mathcal{T} = En_k(F_\mathcal{M} + B_\mathcal{\phi}) = En_k(F_\varphi + 3F_\phi)$.
    \item \textbf{\fedrrr}. The clients need to forward the input images through the feature extractor once. Then, they must compute the matrices $A_k = Z_k^TZ_k$ and $b_k = Z_k^TY_k$. Since $A_k$ is symmetric, computing $A_k$ costs $\frac{1}{2}n_kd(d + 1)$ FLOPs. Instead, computing $b_k$ costs $n_kdC$ FLOPs. Therefore, $\mathcal{T} = n_k (F_\varphi + \frac{1}{2}d(d + 1) + dC)$. 
    \item \textbf{\fedrrrrf}. The costs are the same of \fedrrr, with the sole exception that the latent feature space is $D$-dimensional here. 
    \item \textbf{\fedrrrft}. The costs correspond to those of \fedrrr during the initial phase of the experiments, when \fedrrrft generates the \fedrrr classifier as initialization for the softmax classifier. Subsequently, the costs reflect those of the FT algorithm. For \fedrrrftf, the FT phase costs are $\mathcal{T} = 3En_kF_\mathcal{M}$.
\end{itemize}

\begin{table}[t]

    \caption{MobileNetV2 \cite{mobilenetv2} forward MFLOPs.}
    \label{tab:fw_costs}
    \centering
    \begin{adjustbox}{width=.3\linewidth}
        \centering
        
        \begin{tabular}{lccc}
            
            \toprule

            \textbf{Dataset} & $F_\varphi$ & $F_\phi$ & $F_\mathcal{M}$ \\
            
            \midrule

            \gld & 332.9 & 2.6 & 335.5 \\
            \inat & 332.9 & 1.5 & 334.4 \\
            \cifar{} & 332.9 & 0.1 & 333.0 \\
            
            \bottomrule
        
        \end{tabular}
    \end{adjustbox}
    
\end{table}

The specific forward FLOPs of the MobileNetV2 model are summarized in \cref{tab:fw_costs}.

\section{Centralized \rr Results Using the Random Features}
\label{sec:brrrf}

The outcomes of centralized experiments employing random features to approximate the RBF kernel empirically show that augmenting the number of random features significantly enhances performance. Specifically, \cref{fig:centr_brr_all} illustrates how, with the random features approximation, the performance of \rr calculated over the feature maps provided by the feature extractor across the entire \gld dataset eventually approaches the upper bound established by the exact KRR solution on a subset of the dataset, where a maximum of $40$ images per class is considered.

It is noteworthy that the exact KRR solution was not computed over the entire dataset due to computational constraints. Indeed, the exact solution would require storing a kernel matrix of dimensionality $n \times n$, where $n = 164172$ for \gld. Nevertheless, utilizing the whole dataset or increasing the number of random features should theoretically improve results further.

In addition, \cref{fig:centr_brr_rf_conv} empirically shows that KRR can even yield superior performance compared to a softmax classifier at convergence.

\begin{figure}
\centering
\begin{subfigure}[b]{0.48\textwidth}
    \centering
    \includegraphics[width=\textwidth]{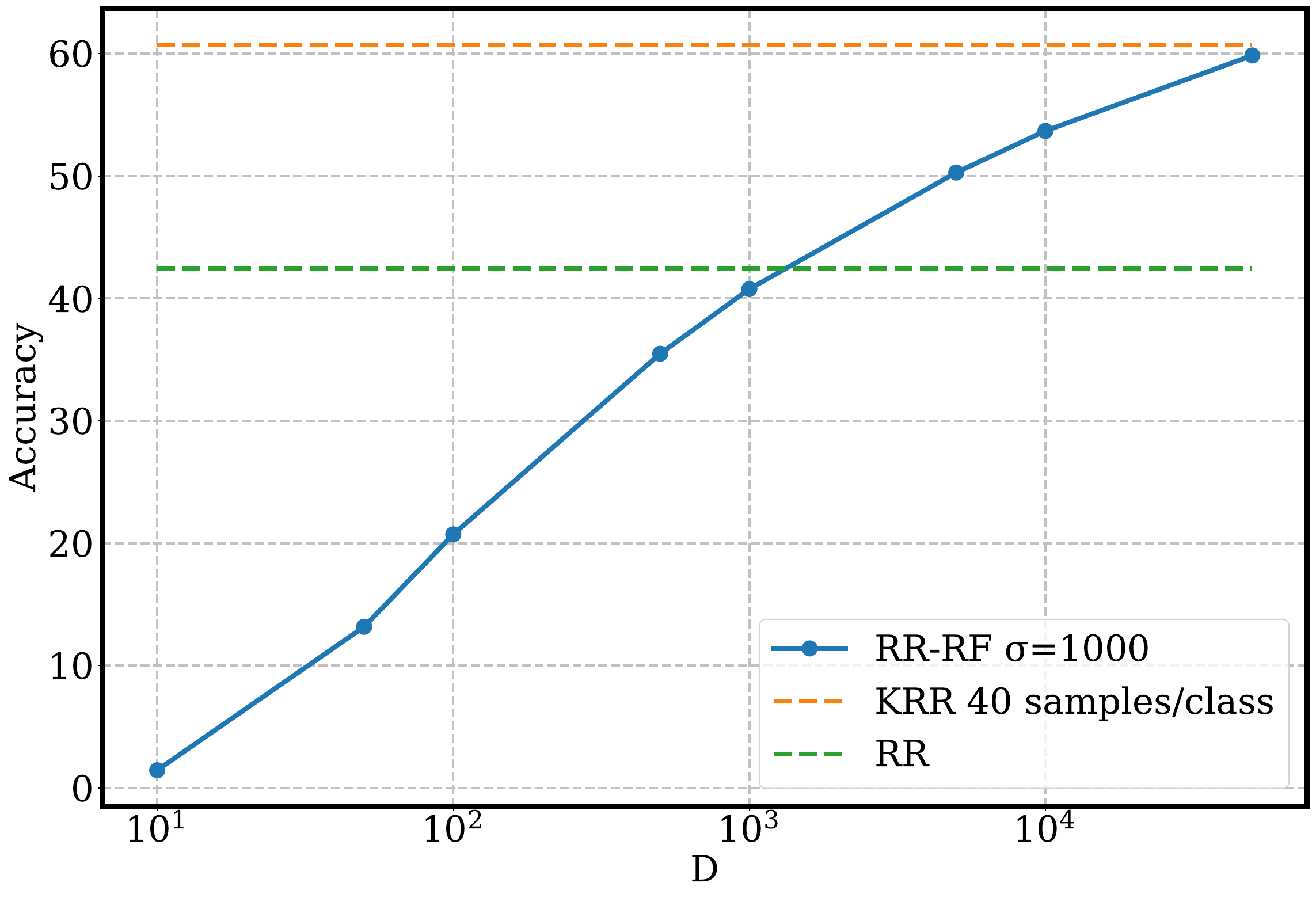}
    \caption{Pre-trained on ImageNet}
    \label{fig:centr_brr_rf}
\end{subfigure}
\quad
\begin{subfigure}[b]{0.48\textwidth}
    \centering
    \includegraphics[width=\textwidth]{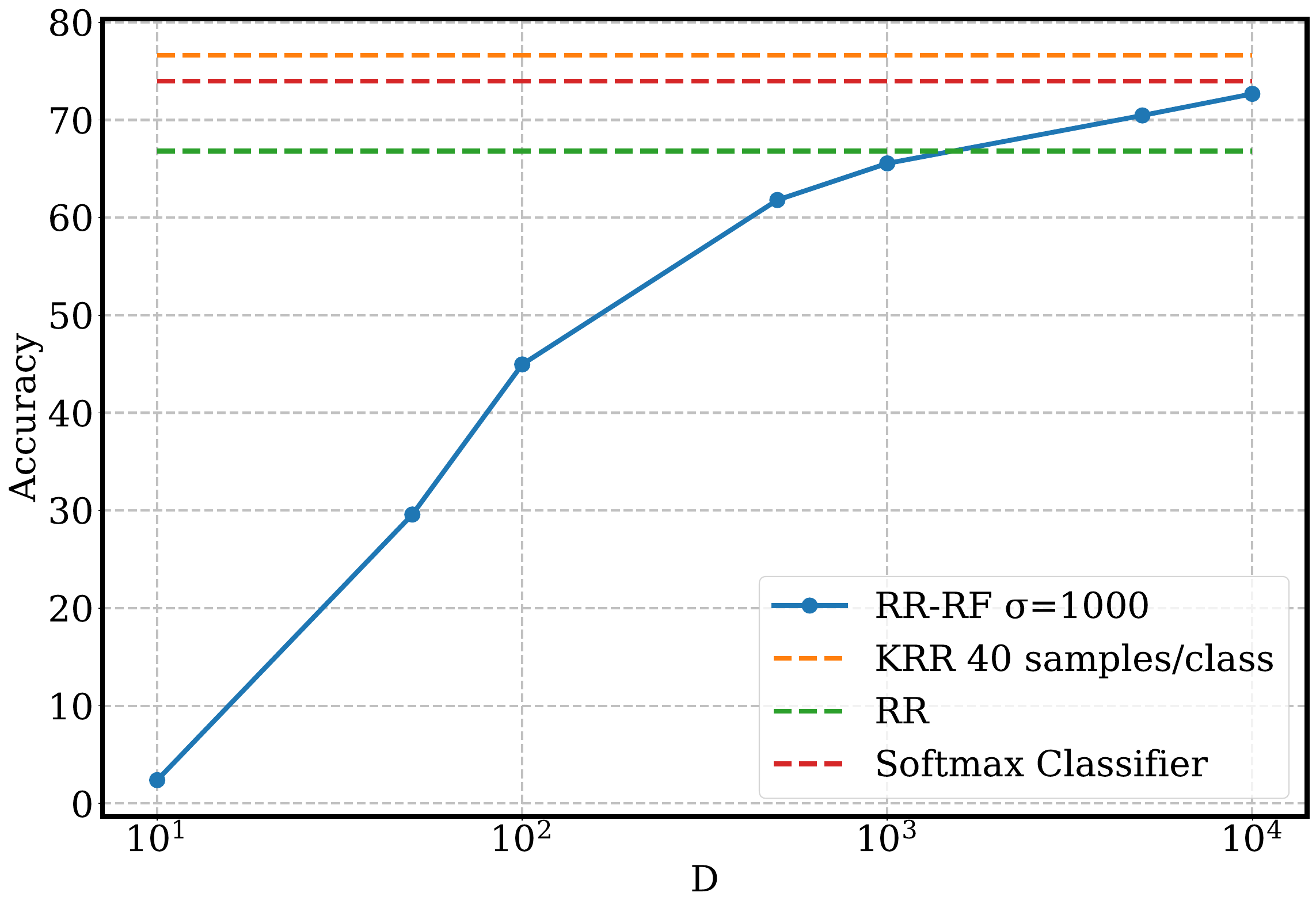}
    \caption{Fine-tuned on Landmarks-Users-160K}
    \label{fig:centr_brr_rf_conv}
\end{subfigure}
\caption{(Centralized) \rr using $D$ random features to approximate the RBF kernel compared to the exact KRR solution with RBF kernel computed over a subset of the whole Landmarks-Users-160K dataset where there are at most $40$ images per class, using the MobileNetV2 \cite{mobilenetv2} architecture. We keep $\sigma = 1000$ for both KRR and \rr-RF.}
\label{fig:centr_brr_all}
\end{figure}

\section{Cifar100 Experiments}
\label{sec:cifar}

\begin{figure}
    \centering
    \includegraphics[width=.5\linewidth]{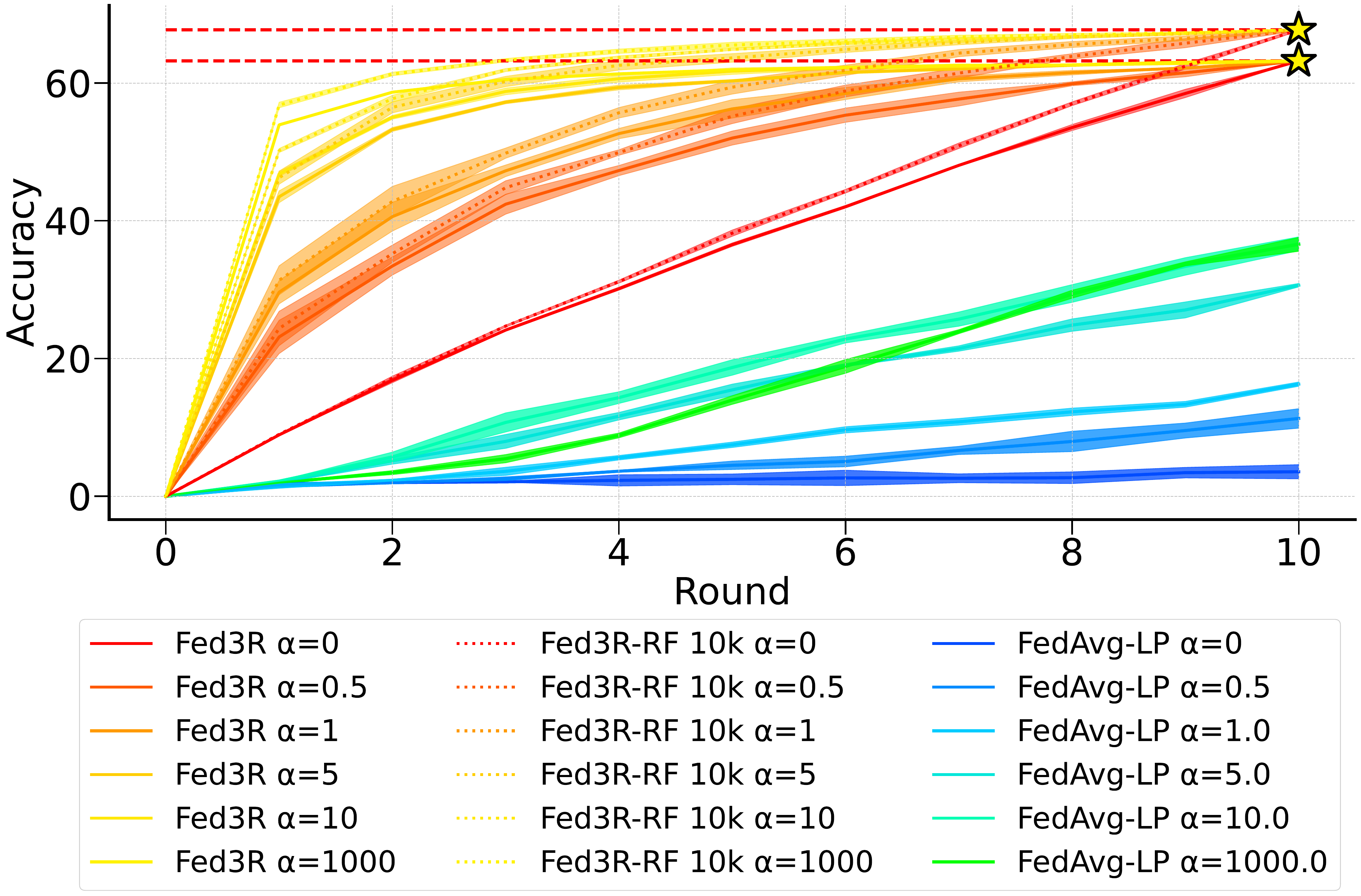}
    \caption{\fedrrr and \fedrrrrf immunity to statistical heterogeneity showed with several \cifar{} FL splits. A lower value of $\alpha$ is associated with a higher level of statistical heterogeneity.}
    \label{fig:cifar-sh-immunity}
\end{figure}

\begin{table}[t]

    \caption{Final accuracy ($\%$) of the \fedrrr family of classifiers and FedNCM on the \cifar{} dataset.}
    \label{tab:fedncm-cifar}
    \centering
    \begin{adjustbox}{width=.3\linewidth}
        \centering
        
        \begin{tabular}{lc}
            
            \toprule

            \textbf{Algorithm} & \textbf{Accuracy ($\%$)} \\
            
            \midrule

            \fedrrr & 63.2 \\
            \fedrrrrf 5k & 66.2 \\
            \textbf{\fedrrrrf 10k} & \textbf{67.5} \\
            \fedncm & 51.0 \\
            
            \bottomrule
        
        \end{tabular}
    \end{adjustbox}
    
\end{table}

\cref{fig:cifar-sh-immunity} shows the immunity to statistical heterogeneity property of \fedrrr and \fedrrrrf also for several \cifar{} FL splits. The number of rounds necessary to converge is only 10 since there are only 100 clients, and we simulate a participation rate of 0.1 \ie we sample 10 clients per round. Moreover, \cref{tab:fedncm-cifar} shows a comparison between the \fedrrr and \fedrrrrf classifiers with the \fedncm classifier \cite{legate2023guiding}.

\begin{figure}[H]
    \centering
    \begin{subfigure}[b]{0.33\linewidth}
        \includegraphics[width=\linewidth]{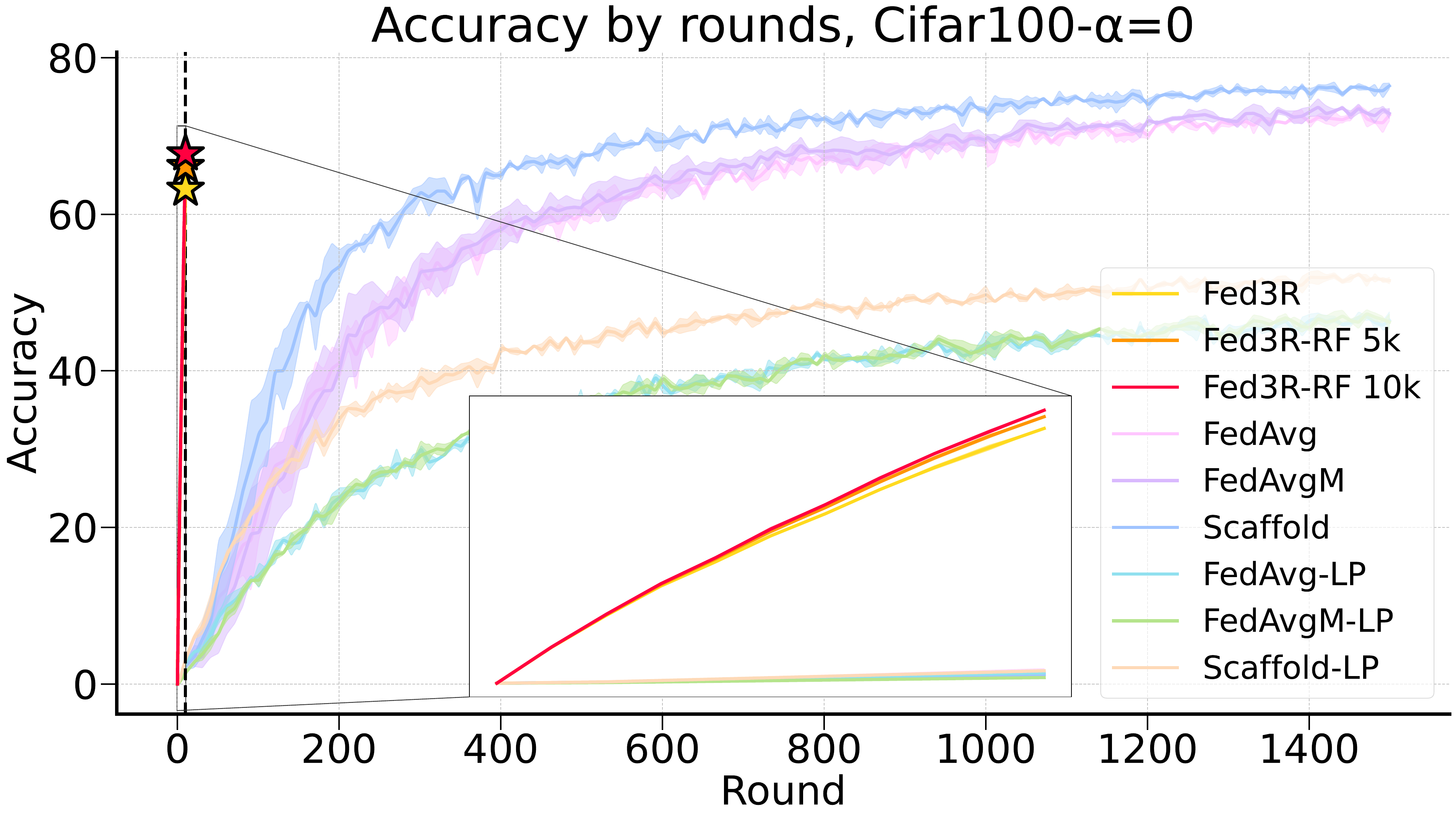}
    \end{subfigure}
    \begin{subfigure}[b]{0.33\linewidth}
        \includegraphics[width=\linewidth]{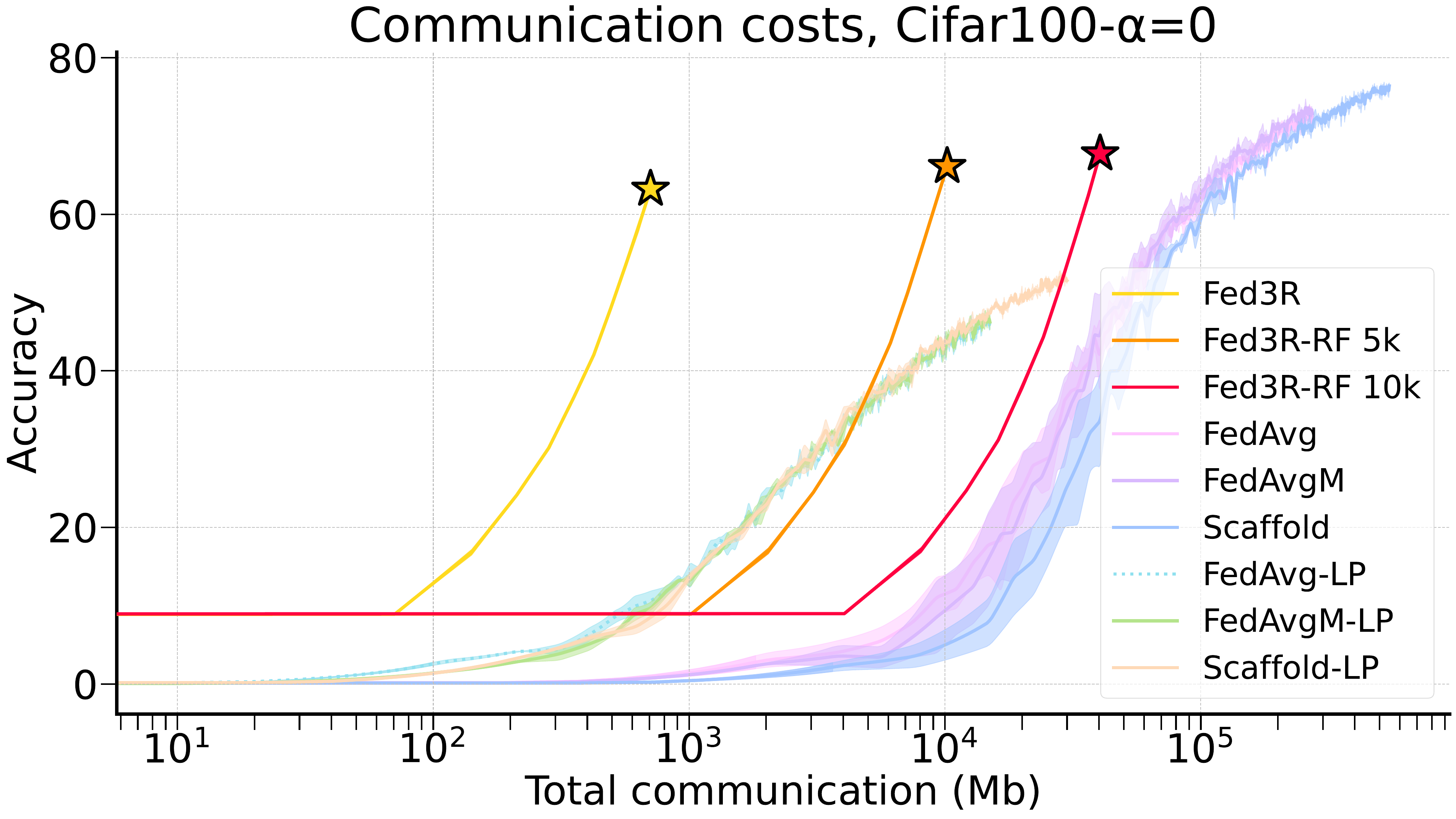}
    \end{subfigure}
    \begin{subfigure}[b]{0.33\linewidth}
        \includegraphics[width=\linewidth]{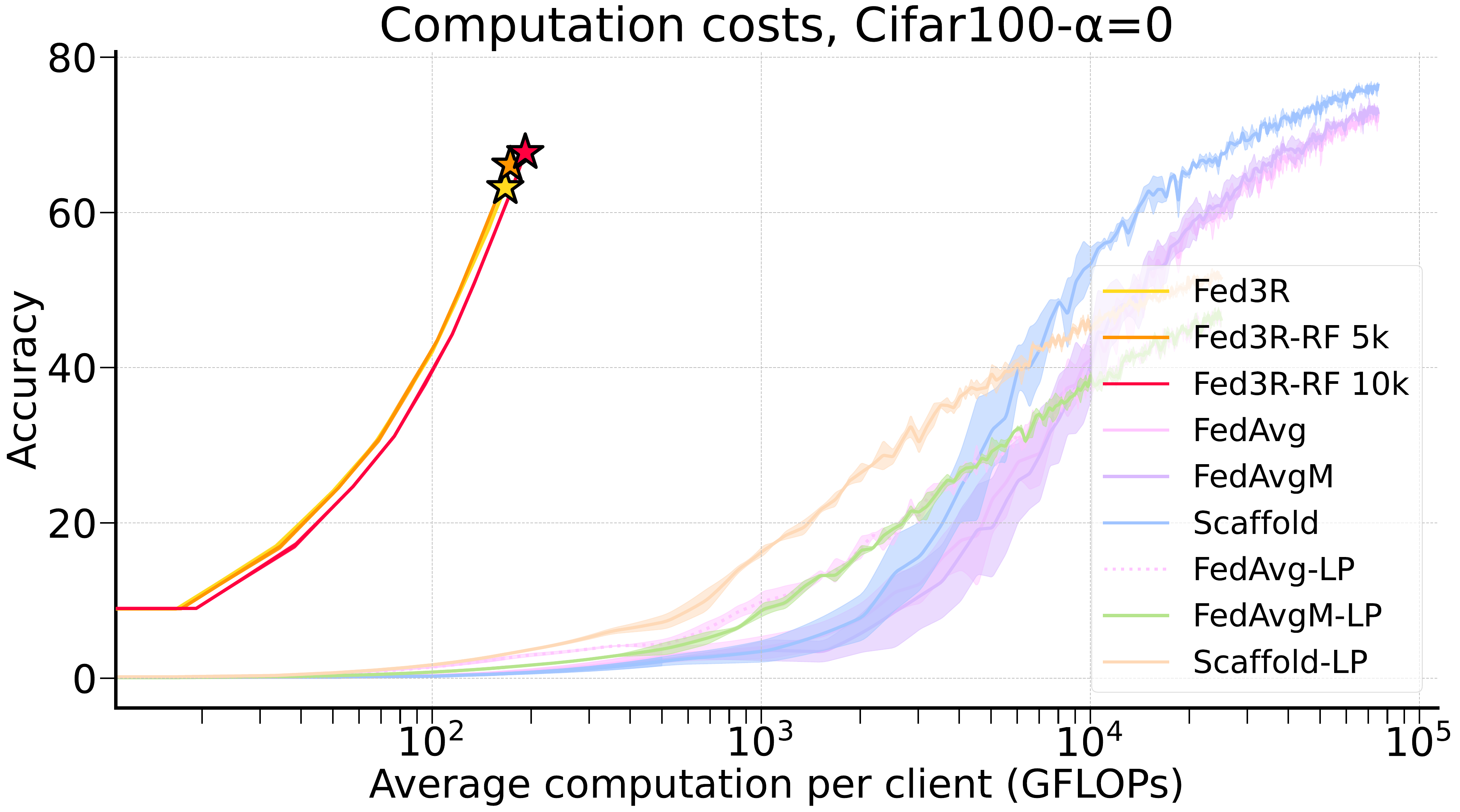}
    \end{subfigure}
    \caption{Comparison between \fedrrr and the baselines for the \cifar{0} dataset.}
    \label{fig:fed3r-cifar}
\end{figure}

\begin{figure}[H]
    \centering
    \begin{subfigure}[b]{0.33\linewidth}
        \includegraphics[width=\linewidth]{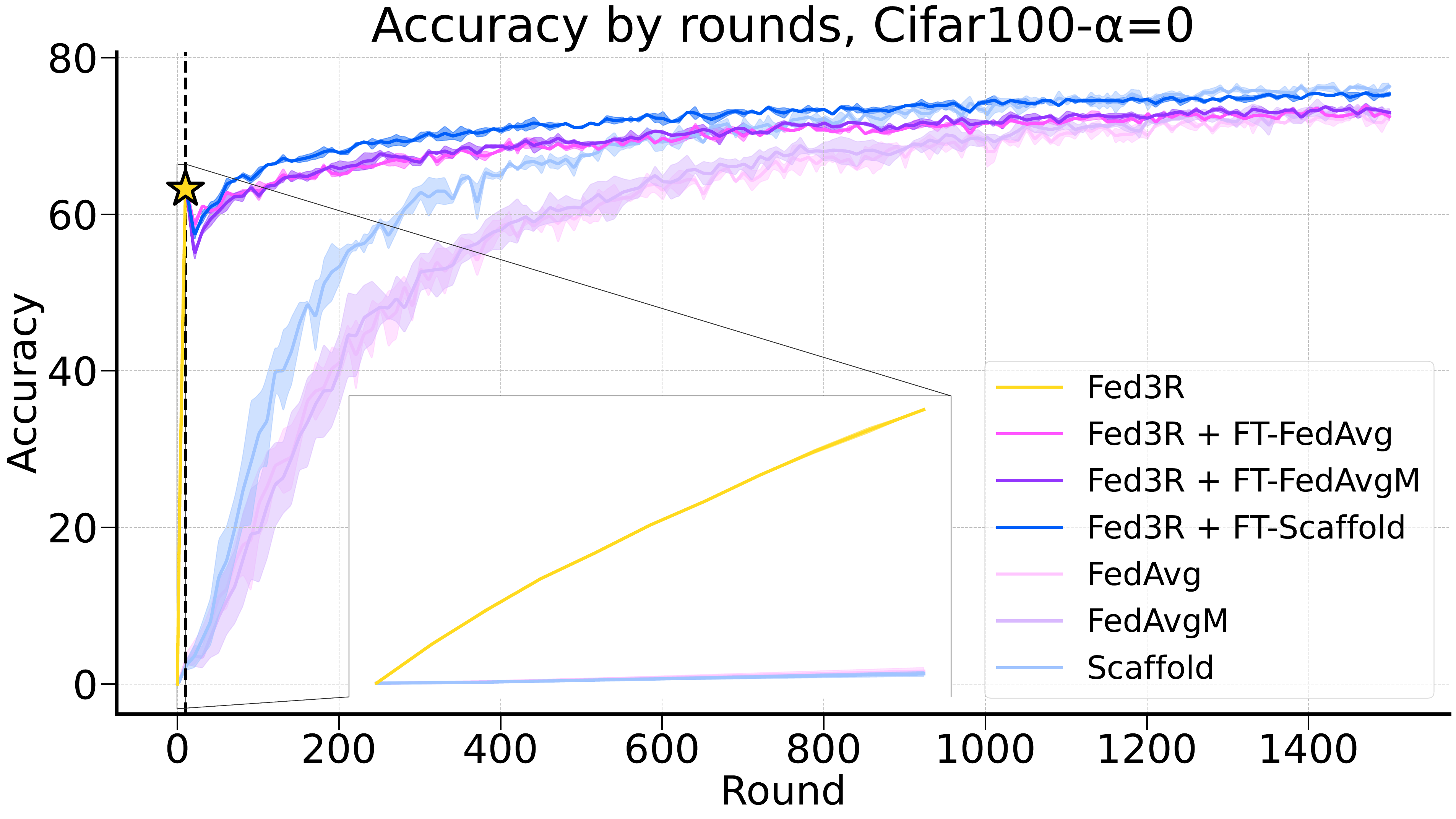}
    \end{subfigure}
    \begin{subfigure}[b]{0.33\linewidth}
        \includegraphics[width=\linewidth]{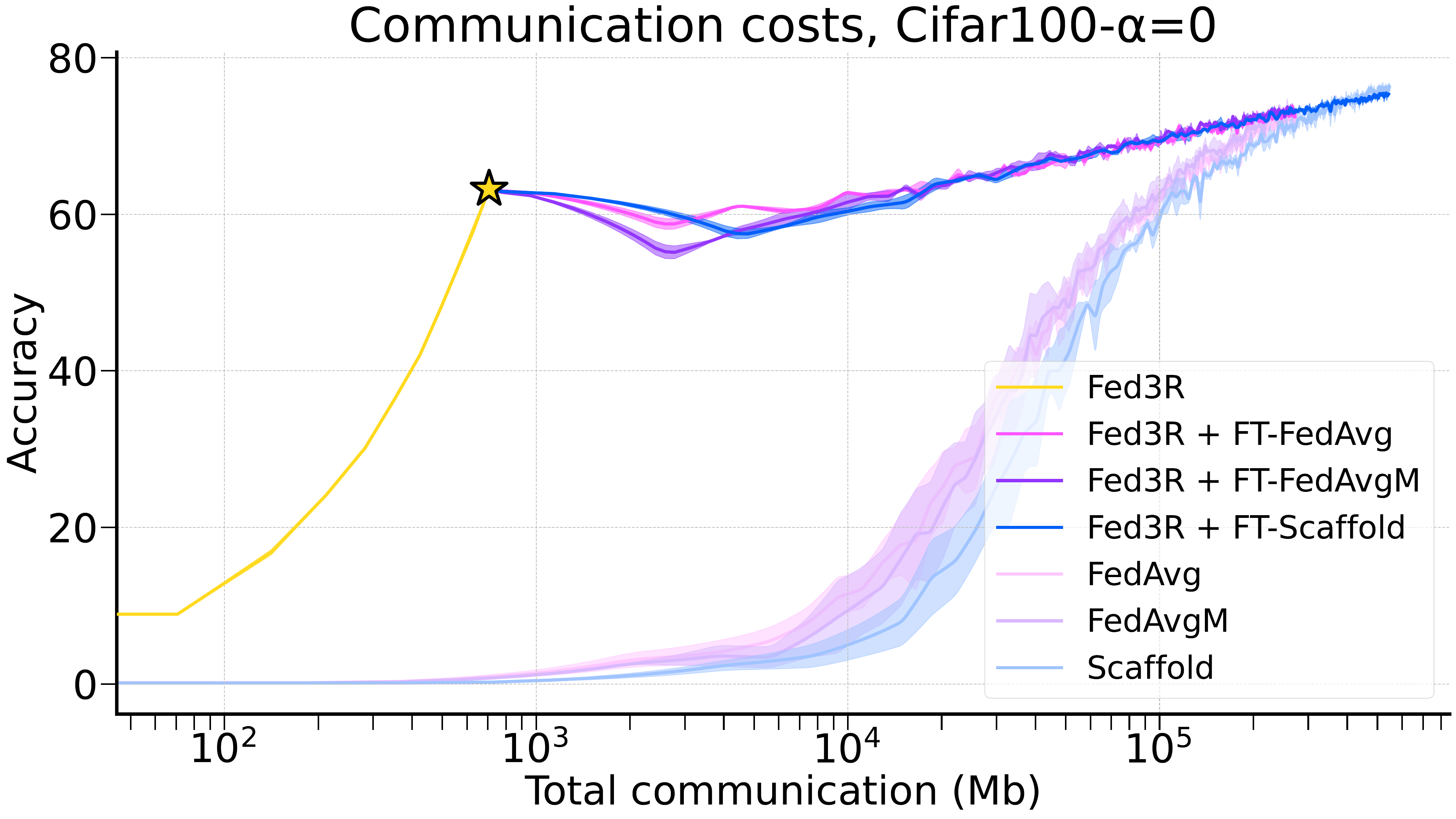}
    \end{subfigure}
    \begin{subfigure}[b]{0.33\linewidth}
        \includegraphics[width=\linewidth]{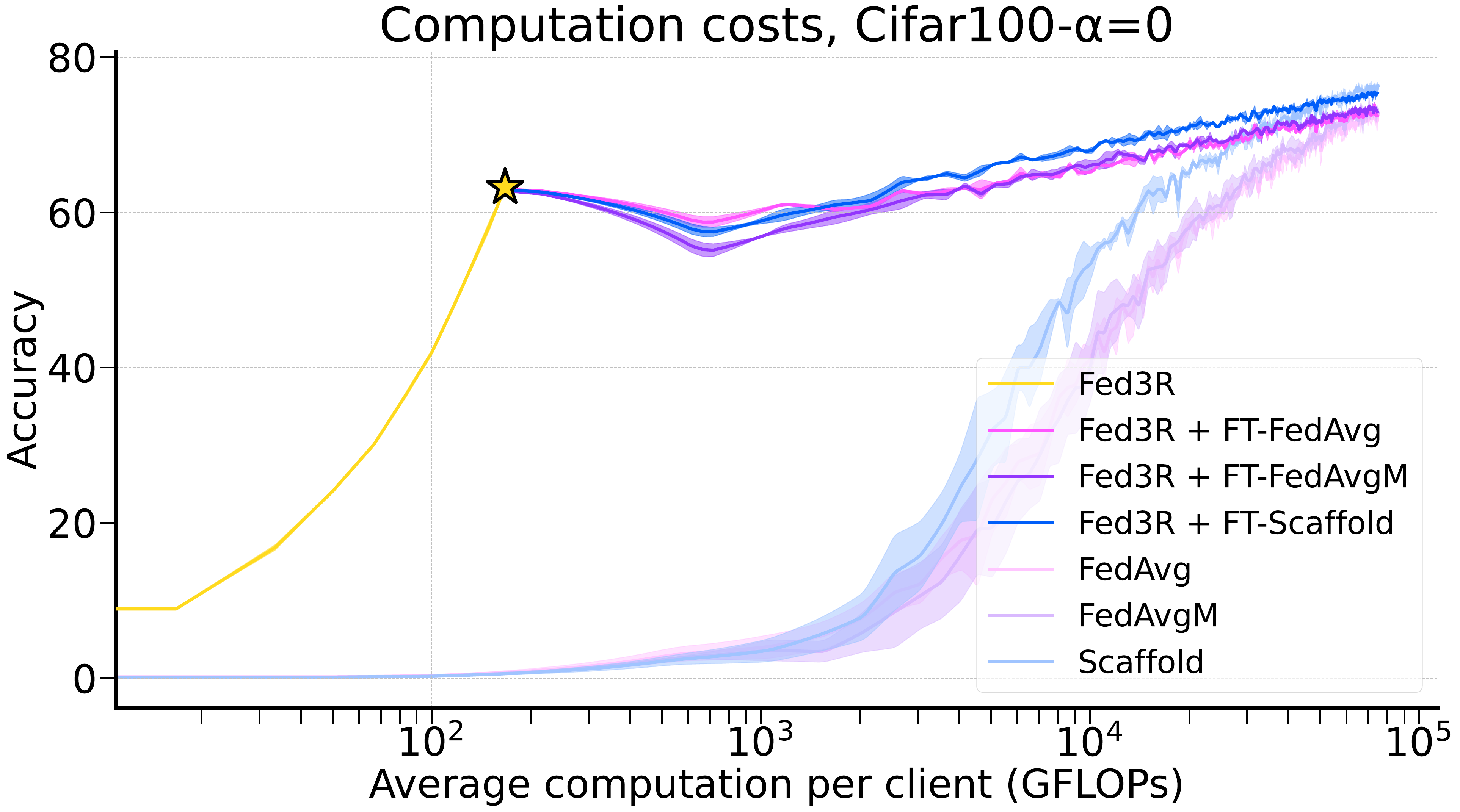}
    \end{subfigure}
    \caption{Comparison between \fedrrrft and the baselines for the \cifar{0} dataset.}
    \label{fig:fed3rft-cifar}
\end{figure}

Furthermore, \cref{fig:fed3r-cifar} illustrates the performance and costs of both \fedrrr and \fedrrrrf, whereas \cref{fig:fed3rft-cifar} showcases the performance and costs of \fedrrrft. While \fedrrr alone may not achieve satisfactory performance compared to the baselines in this scenario, incorporating the fine-tuning stage remarkably enhances its performance to a comparable or superior level with respect to the baselines while retaining the same communication and computation advantages.

\section{Best Methods Comparison}
\label{sec:best}

\cref{fig:bestgld,fig:bestinat} show a comparison among the best \fedrrr algorithms and the best baseline for the \gld and \inat datasets, respectively.

\fedrrrft provides the best performance for \gld, and has better communication costs than \fedrrrrf at the maximum \fedrrrrf accuracy ($56.6\%$). However, if the computation budget is the main constraint, \fedrrrrf is the best method for computation budget $\leq 10^3$ GFLOPs. Similarly, \fedrrrrf is the best method up to $10^2$ GFLOPs per client in the \inat experiments, although suffering high communication costs.

In all the cases, all our best-performing methods are much better than the best of the baselines, with the sole exception of the communication costs required to reach a target accuracy of \fedrrrrf, which are comparable with the best baseline in the final stages of the training.

\begin{figure}[H]
    \centering
    \begin{subfigure}[b]{0.33\linewidth}
        \includegraphics[width=\linewidth]{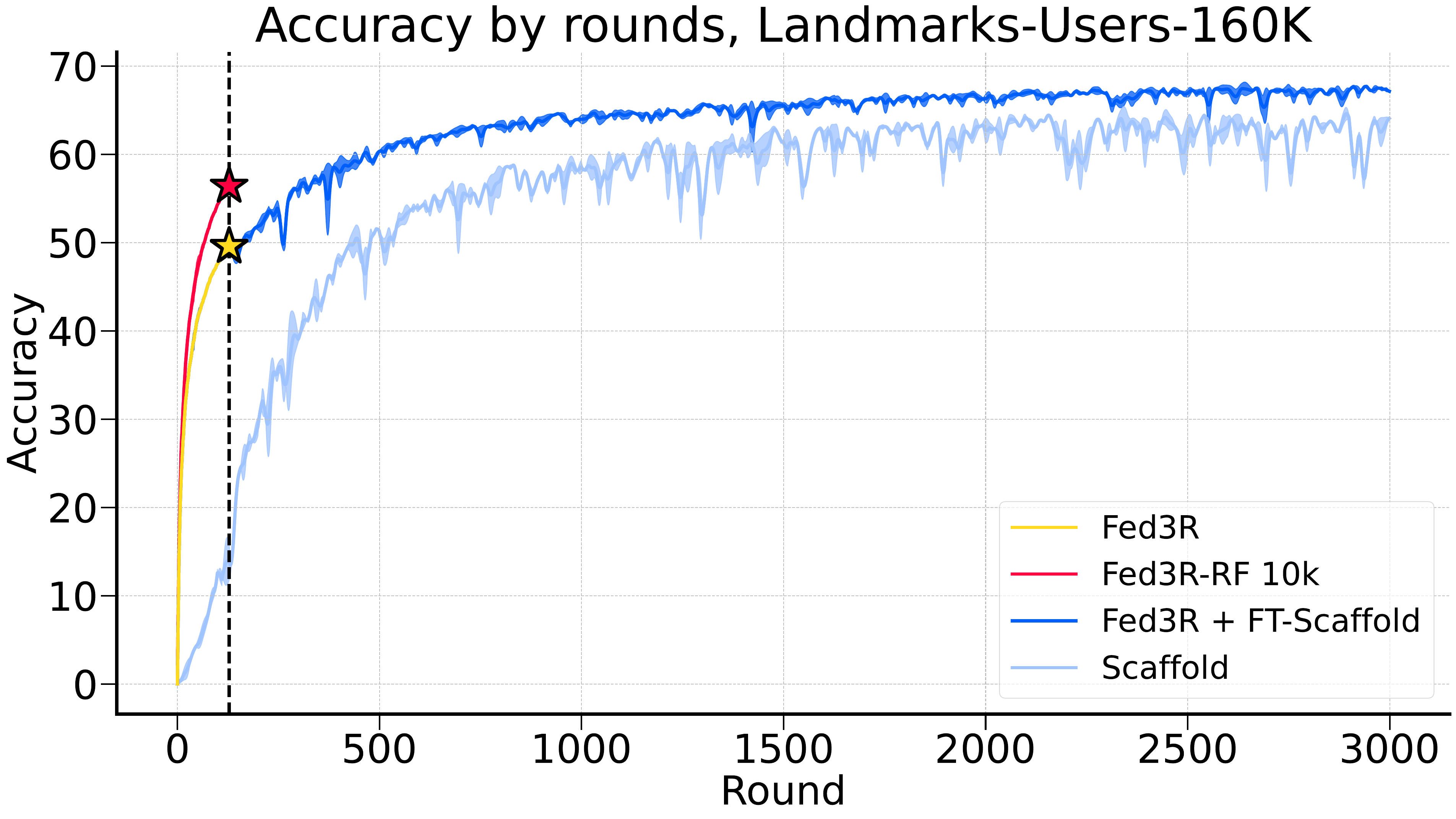}
    \end{subfigure}
    \begin{subfigure}[b]{0.33\linewidth}
        \includegraphics[width=\linewidth]{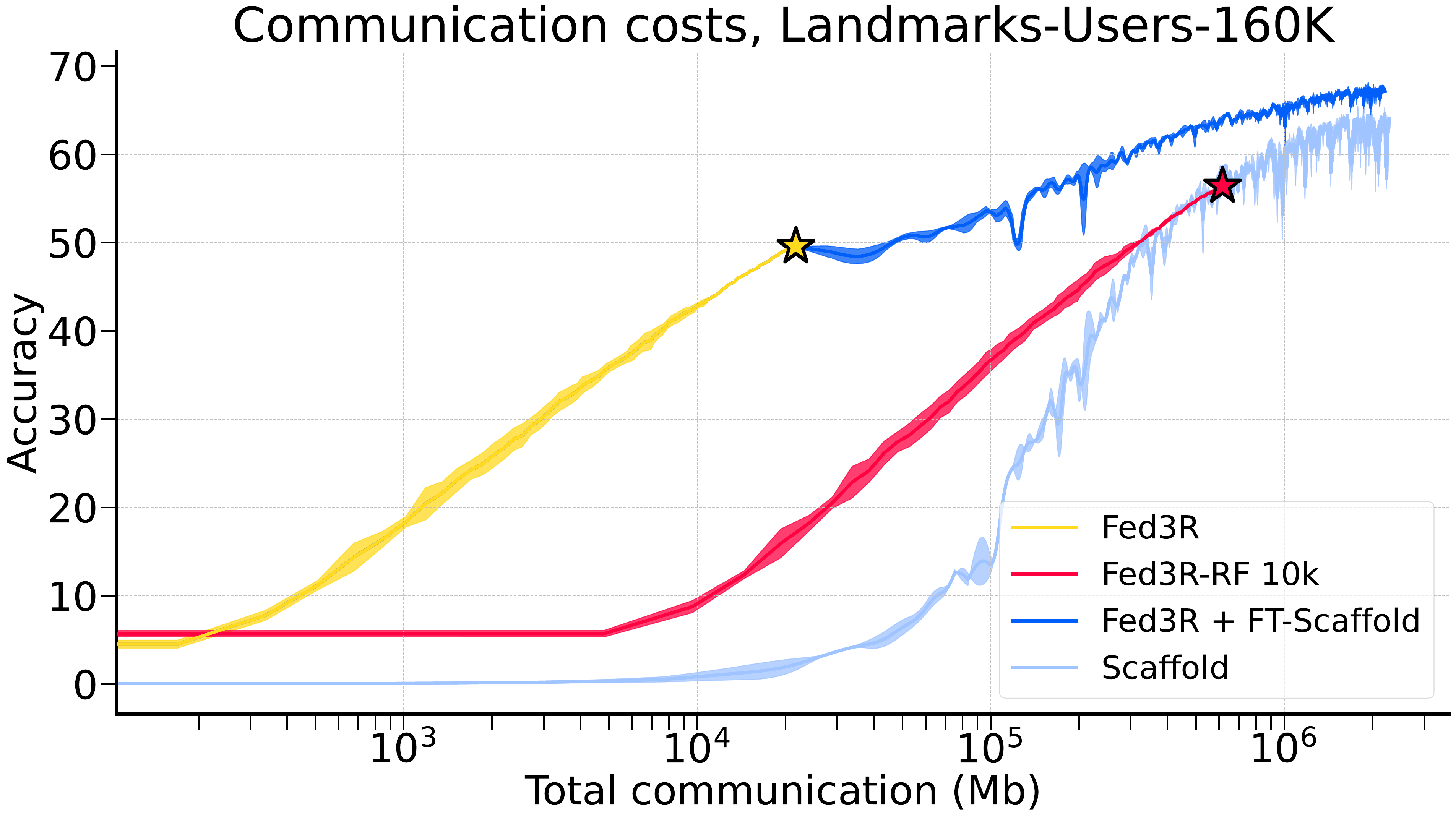}
    \end{subfigure}
    \begin{subfigure}[b]{0.33\linewidth}
        \includegraphics[width=\linewidth]{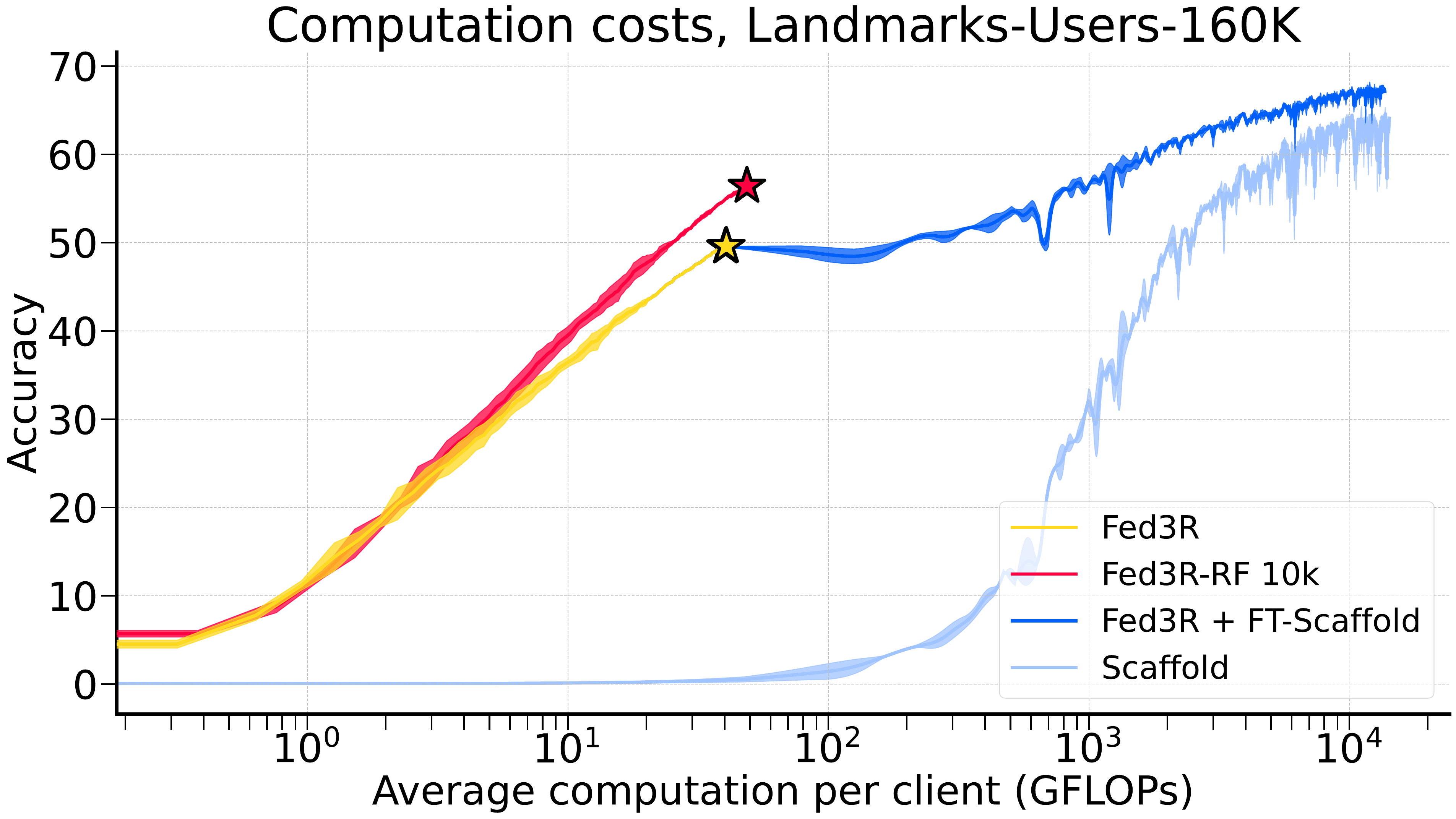}
    \end{subfigure}
    \caption{Comparison between the best \fedrrr methods and the best baseline method for the \gld dataset.}
    \label{fig:bestgld}
\end{figure}

\begin{figure}[H]
    \centering
    \begin{subfigure}[b]{0.33\linewidth}
        \includegraphics[width=\linewidth]{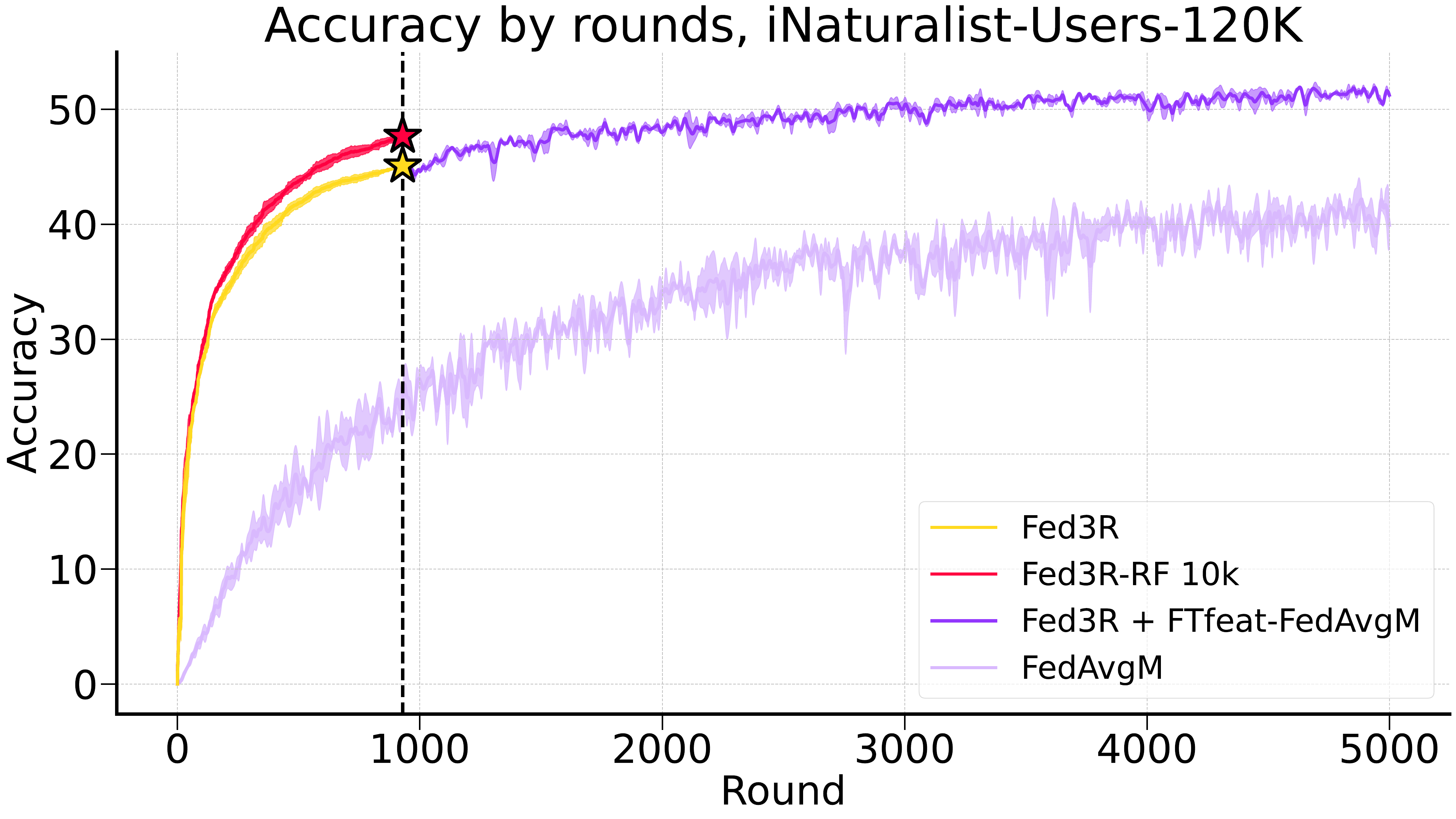}
    \end{subfigure}
    \begin{subfigure}[b]{0.33\linewidth}
        \includegraphics[width=\linewidth]{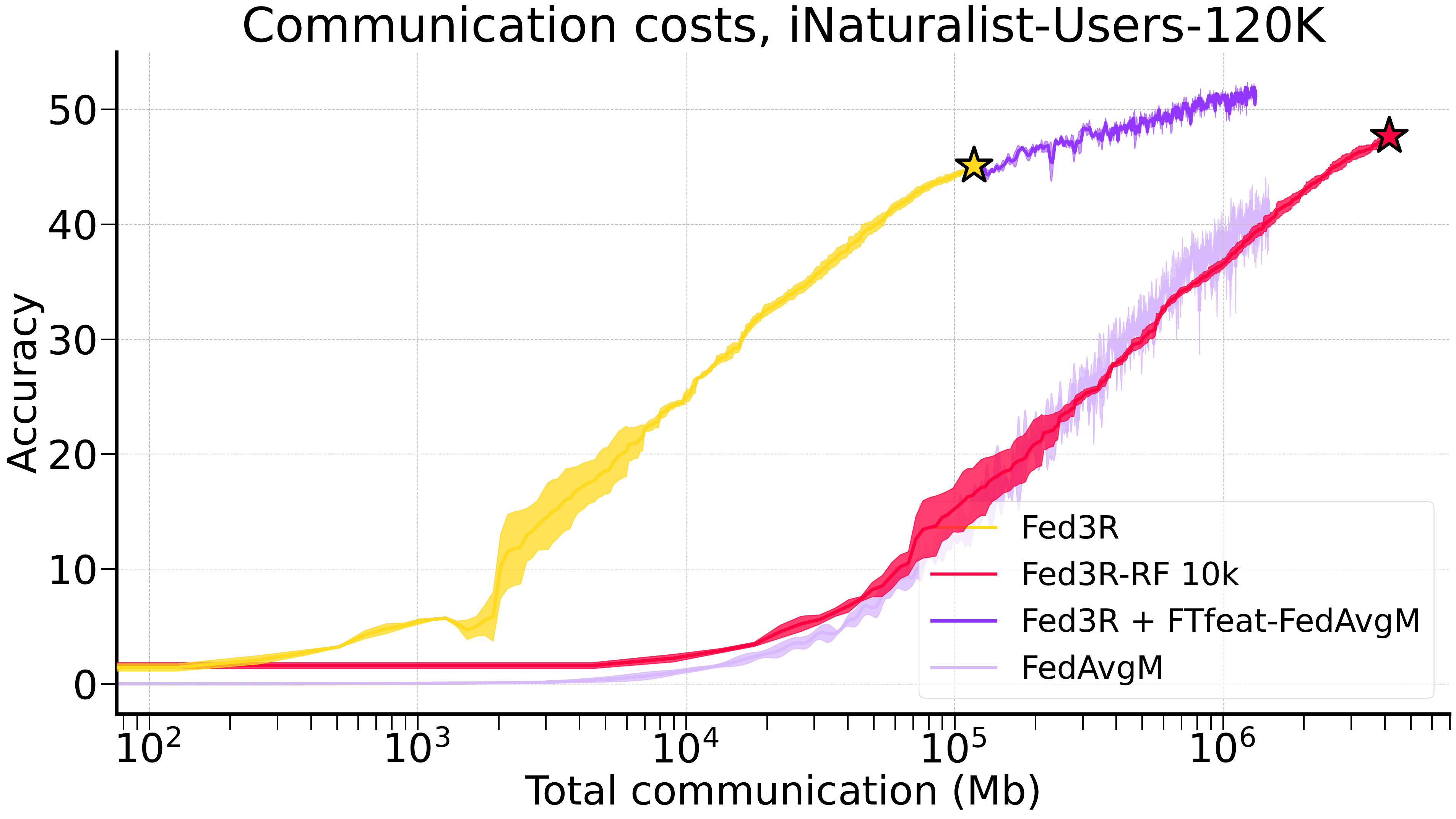}
    \end{subfigure}
    \begin{subfigure}[b]{0.33\linewidth}
        \includegraphics[width=\linewidth]{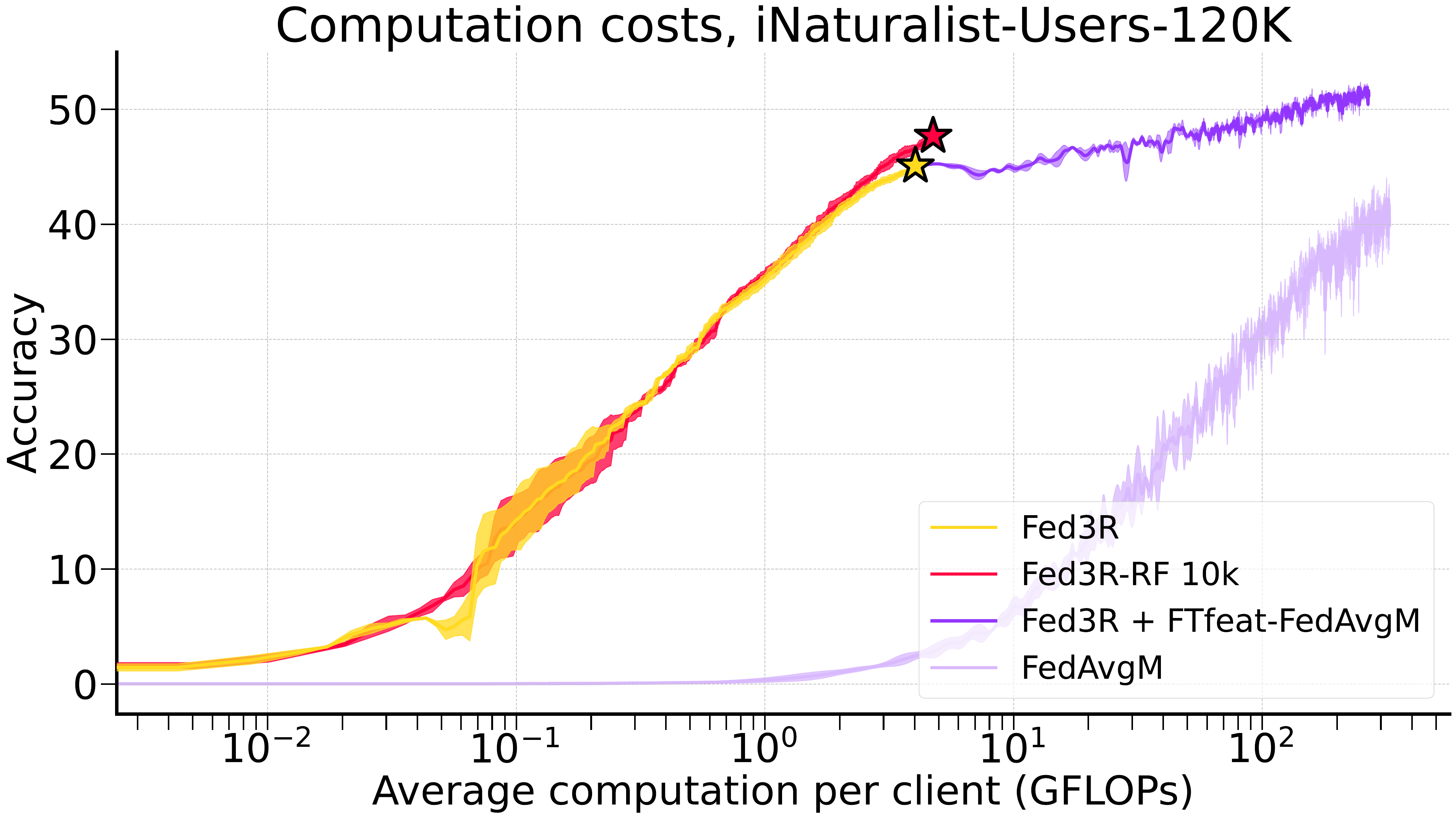}
    \end{subfigure}
    \caption{Comparison between the best \fedrrr methods and the best baseline method for the \inat dataset.}
    \label{fig:bestinat}
\end{figure}

\section{Expected Number of Rounds to Sample Each Client at Least Once With Replacement}
\label{sec:coupons}

Given the total number of clients $K$ in the federation and the number of clients $\kappa$ sampled per round, it is theoretically possible to estimate the expected number of rounds that are necessary to sample each client at least once. This problem is known in the literature as the Batch Coupon Collector's Problem \cite{coupon0, coupon1, coupon}.

\cref{tab:coupon} shows the average number of rounds necessary to sample a given percentage of clients in the corresponding settings after simulating the sampling with replacement one thousand times for each case. It is possible to observe how, to sample all the clients, many more rounds are needed on average than the rounds needed to sample 50$\%$ or 75$\%$ distinct clients. Interestingly, regarding the scenarios involving sampling with replacement as the one in \cref{fig:suboptimal}, this table provides insight into why \fedrrr achieves good performance with few rounds compared to the total needed to achieve full convergence.

\begin{table}[H]

    \caption{Average number of rounds necessary to sample the given percentage of clients in the corresponding settings when the clients are sampled with replacement.}
    \label{tab:coupon}
    \centering
    \begin{adjustbox}{width=.7\linewidth}
        \centering
        
        \begin{tabular}{lccccccc}
            
            \toprule

            \textbf{Dataset} & $K$ & $\kappa$ & Participation Rate ($\%$) & $25\%$ & $50\%$ & $75\%$ & $100\%$ \\
            
            \midrule

            \multirow{3}{*}{\gld} & \multirow{3}{*}{1262} & 10 & 0.8 & 37 \tiny{$\pm$ 1} & 88 \tiny{$\pm$ 2} & 175 \tiny{$\pm$ 4} & 970 \tiny{$\pm$ 155} \\
             & & 20 & 1.6 & 19 \tiny{$\pm$ 1} & 44 \tiny{$\pm$ 1} & 87 \tiny{$\pm$ 2} & 483 \tiny{$\pm$ 79} \\
             & & 50 & 4.0 & 8 \tiny{$\pm$ 0} & 18 \tiny{$\pm$ 1} & 35 \tiny{$\pm$ 1} & 191 \tiny{$\pm$ 32} \\

            \midrule
             
            \multirow{3}{*}{\inat} & \multirow{3}{*}{9275} & 10 & 0.1 & 267 \tiny{$\pm$ 2} & 643 \tiny{$\pm$ 5} & 1286 \tiny{$\pm$ 12} & 9020 \tiny{$\pm$ 1189} \\
             & & 20 & 0.2 & 134 \tiny{$\pm$ 1} & 322 \tiny{$\pm$ 3} & 643 \tiny{$\pm$ 6} & 4494 \tiny{$\pm$ 596} \\
             & & 50 & 0.5 & 54 \tiny{$\pm$ 1} & 129 \tiny{$\pm$ 1} & 257 \tiny{$\pm$ 2} & 1809 \tiny{$\pm$ 247} \\

            \midrule
             
            \multirow{3}{*}{\cifar{}} & \multirow{3}{*}{100} & 10 & 10 & 3 \tiny{$\pm$ 0} & 7 \tiny{$\pm$ 1} & 14 \tiny{$\pm$ 1} & 50 \tiny{$\pm$ 12} \\
             & & 20 & 20 & 2 \tiny{$\pm$ 0} & 4 \tiny{$\pm$ 0} & 7 \tiny{$\pm$ 1} & 24 \tiny{$\pm$ 5} \\
             & & 50 & 50 & 1 \tiny{$\pm$ 0} & 1 \tiny{$\pm$ 0} & 3 \tiny{$\pm$ 0} & 8 \tiny{$\pm$ 2} \\
            
            \bottomrule
        
        \end{tabular}
    \end{adjustbox}
    
\end{table}

\end{document}